%% file: Manuscript.tex
\documentclass[a4paper,fleqn]{cas-dc}

\usepackage[numbers]{natbib}

\usepackage{graphicx}%
\usepackage{multirow}%
\usepackage{amsmath,amssymb,amsfonts}%
\usepackage{amsthm}%
\usepackage{mathrsfs}%
\usepackage[title]{appendix}%
\usepackage{xcolor}%
\usepackage{textcomp}%
\usepackage{manyfoot}%
\usepackage{booktabs}%
\usepackage{algorithm}%
\usepackage{algorithmicx}%
\usepackage{algpseudocode}%
\usepackage{listings}%

\usepackage{orcidlink}
\usepackage{forest}
\usepackage{tabularx}
\usepackage{pifont} 
\usepackage{makecell}
\usepackage{ragged2e}  
\usepackage{rotating} 
\usepackage{hyperref}
\usepackage{threeparttablex}

\begin{document}
\let\WriteBookmarks\relax
\def\floatpagepagefraction{1}
\def\textpagefraction{.001}
\shorttitle{Out-of-Distribution Generalization in Time Series: A Survey}
\shortauthors{X. Wu et~al.}
      
\title [mode = title]{Out-of-Distribution Generalization in Time Series: A Survey}                      



\author[1]{Xin Wu}[style=chinese,orcid=00009-0008-2292-4534]


\affiliation[1]{organization={School of Computing and Artificial Intelligence, Southwest Jiaotong University},
                city={Chengdu},
                postcode={695013}, 
                country={China}}

\author[1,2]{Fei Teng}[style=chinese,orcid=0000-0001-9535-7245]
\cormark[1]
\ead{fteng@swjtu.edu.cn}

\author[1]{Xingwang Li}[style=chinese,orcid=0009-0007-9650-130X]

\author[1]{Ji Zhang}[style=chinese,orcid=0000-0001-6949-3673]

\author[3]{Qiang Duan}[style=chinese,orcid=0000-0001-7832-1937]

\author[1]{Tianrui Li}[style=chinese,orcid=0000-0001-7780-104X]

\affiliation[2]{organization={Engineering Research Center of Sustainable Urban Intelligent Transportation, Ministry of Education},
                city={Chengdu},
                postcode={611756}, 
                country={China}}

\affiliation[3]{organization={Information Sciences and Technology Department, the Pennsylvania State University},
                city={Abington},
                postcode={19001}, 
                state={PA}, 
                country={USA}}

\cortext[cor1]{Corresponding author.}


\begin{abstract}
Time series frequently manifest distribution shifts, diverse latent features, and non-stationary learning dynamics, particularly in open and evolving environments. These characteristics pose significant challenges for out-of-distribution (OOD) generalization. While substantial progress has been made, a systematic synthesis of advancements remains lacking. To address this gap, we present the comprehensive review of OOD generalization methodologies for time series, organized to delineate the field's evolutionary trajectory and contemporary research landscape. We organize our analysis across three foundational dimensions: data distribution, representation learning, and OOD evaluation. For each dimension, we present several popular algorithms in detail. Furthermore, we highlight key application scenarios, emphasizing their real-world impact. Finally, we identify persistent challenges and propose future research directions. A detailed summary of the methods reviewed for the generalization of OOD in time series can be accessed at \textcolor{blue}{\url{https://tsood-generalization.com}}.
\end{abstract}


%
%
%
%
%

%
%
%


\maketitle

\section{Introduction}

\label{sec:introduction}
\input{1_introduction}

\section{Background}
\label{sec-pre}

\input{2_background}

\section{Taxonomy}
\label{sec-method}

\begin{figure*}
	\centering
	\resizebox{\textwidth}{!}{
	\input{fig_tree}
	}
	\caption{A taxonomy examining methods for TS-OOG from three perspectives: data distribution, representation learning, and OOD evaluation. This includes various approaches like covariate shift adaptation, invariant learning, adaptive mechanism-based methods, and large time series models (LTSMs). Note that OOD evaluation specifically for LTSMs is not summarized here due to the current lack of standardized evaluation protocols.}
	\label{fig-main}
\end{figure*}

This section will provide a detailed overview of the TS-OOG methods. As shown in Fig. \ref{fig-main}, we discuss these methods from the following three perspectives:

\begin{enumerate}[(1)]
    \item \emph{Data Distribution:} These methods primarily focus on addressing the differences in data distribution between the training and test datasets. Two popular approaches under this concept are: \emph{a). Covariate Shift:} This assumes that the distribution of input features changes between the training and testing phases, while the conditional distribution \( \mathbb{P} \left (Y \mid X\right )  \)  remains unchanged. \emph{b). Concept Drift:} As time progresses or the environment changes, the conditional distribution \( \mathbb{P} \left (Y \mid X\right )  \)  shifts. In other words, there is a fundamental change in the concept or environment dynamics between the training and testing stages.
    \item \emph{Representation learning:} These methods aim to reduce or eliminate the differences between data distributions by learning effective feature representations, thereby improving the model's generalization ability. Four representative techniques are: 
    \emph{a). Decoupling-based Methods:} This approach decomposes features into task-relevant and irrelevant parts, thereby eliminating the interference of irrelevant features on the model and enhancing its robustness. 
    \emph{b). Invariant-based Methods:} This approach focuses on extracting features that remain invariant across different domains, ensuring that the model can still rely on these stable features for prediction when facing distribution shifts.
    \emph{c). Adaptive Mechanism-based Methods:} This approach equips models with mechanisms to dynamically adjust to evolving data distributions, enabling real-time adaptation and continual learning, rather than relying solely on static representations.
    \emph{d). LTSMs:} This emerging paradigm utilizes models with massive scale, pre-trained on extensive and diverse datasets. Their strong generalization capability stems from learning universal patterns from a wide array of data distributions, which enhances robustness against distribution shifts. This category is further divided into approaches that adapt large language models (LLMs) and those that pre-train foundation models directly on time series data.
    \item \emph{OOD Evaluation:} These methods assess a model's performance on OOD data and generally include two approaches: 
    \emph{a). Extrinsic OOD Evaluation:} This involves constructing or collecting OOD test datasets with real labels (or generating pseudo-labels) to directly measure the model’s prediction or classification performance across different distributions. 
    \emph{b). Intrinsic OOD Evaluation:} This involves analyzing the internal characteristics of the model (such as uncertainty quantification, consistency of feature distributions, and smoothness of responses) to indirectly infer the model's generalization ability on OOD data.
\end{enumerate}

These three methods are conceptually distinct and can be used individually or in combination to achieve better performance. We will now elaborate on each method in detail.

\section{Data Distribution}
\label{sec4_data}
\input{4_sec_data_distribution}

\section{Representation Learning}
\label{sec5_rep}
\input{5_sec_representation_learning}

\section{OOD Evaluation}
\label{sec6_ood-evaluation}
\input{6_sec_ood_evaluation}

\section{Applications}
\label{sec7_applications}
\input{7_sec_applications}

\section{Challenges and Future Directions}
\label{sec8_future}
\input{8_sec_future}

\section{Conclusion}
\label{sec9_con}
This paper provides a comprehensive review of the current state of research on TS-OOG. From data distribution and representation learning to OOD evaluation, we have summarized the fundamental ideas, technical implementations, and applications of key methodologies while examining the challenges and future directions in this area. Overall, developing robust invariant representations, unified multimodal large-scale models, and robust OOD evaluation frameworks will be instrumental in advancing this field. This survey will offer systematic theoretical guidance and practical references for researchers, facilitate the evolution of time series analysis toward more robust and interpretable next-generation systems, and ultimately enable the construction of highly generalizable time series models under complex real-world conditions.

\bibliographystyle{cas-model2-names}

\bibliography{tsood}


%
%

\end{document}

%% file: 1_introduction.tex
Time series data, sequences of data points ordered chronologically, constitute the cornerstone of numerous real-world applications, spanning domains such as financial analysis \cite{1,2}, autonomous driving \cite{3,4}, healthcare \cite{5,6,fan2025medvia,qin2020imaging,217,218}, and industrial fault diagnosis \cite{7,8,221}. The success of traditional machine learning models is often predicated on the assumption of independent and identically distributed (IID) data, which posits that training and test data are drawn from the same static probability distribution \cite{9,ferrari2014information}. This fundamental premise, however, is fundamentally challenged in the time series domain. Time series are intrinsically characterized by temporal dependency, where the sequence of data points is paramount, and non-stationarity, where statistical properties like mean and variance evolve over time. These characteristics violate the IID assumption, meaning that models trained on historical data are prone to significant performance degradation when deployed on future data from a shifted distribution. Consequently, enabling models to generalize to such out-of-distribution (OOD) data has emerged as a critical scientific problem \cite{liang2025diffusion,wang2025towards,zhao2025neuralood}.

\begin{figure*}
	\centering
	 \includegraphics[width=0.99\textwidth]{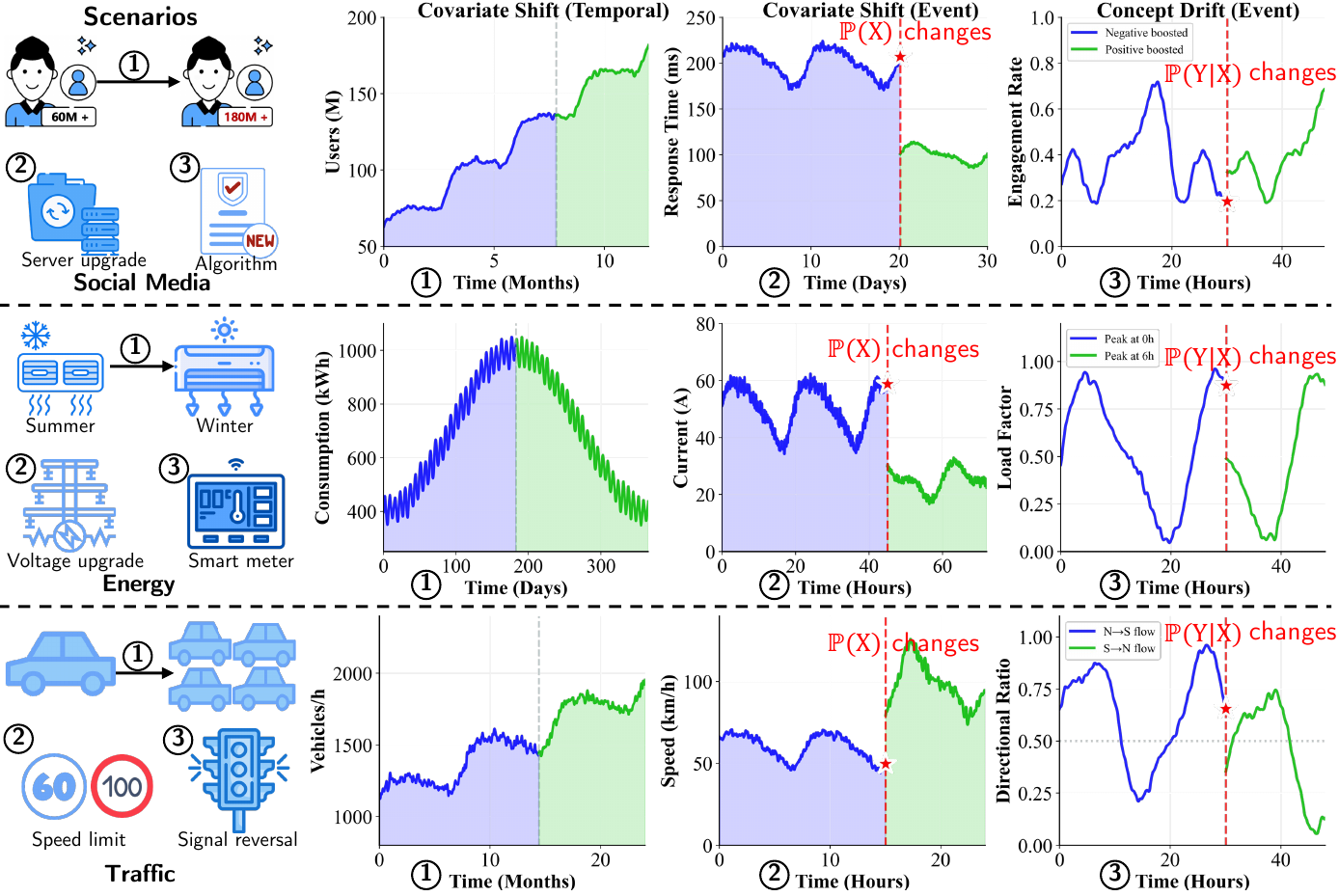}
	\caption{Examples illustrating how real-world temporal dynamics drive distribution shifts in TS-OOG. The first two columns show \textbf{covariate shift}, where the input feature distribution ($\mathbb{P}(X)$) changes due to underlying dynamics like gradual temporal drifts (e.g., user growth) or abrupt events (e.g., server upgrade). The third column depicts \textbf{concept drift}, a more fundamental change where the input-output relationship ($\mathbb{P}(Y|X)$) itself is altered (e.g., by an algorithm change affecting user engagement). In the social media example, both user growth (temporal drift) and server upgrade (event) cause covariate shifts affecting the input data distribution.  Subsequently, the algorithm change induces concept drift by fundamentally altering the relationship between content and engagement. Blue-shaded areas represent historical data; green areas show subsequent OOD data. This highlights that TS-OOG problems stem from temporal dynamics altering either the data's characteristics ($\mathbb{P}(X)$) or its underlying generative rules ($\mathbb{P}(Y|X)$).}
	\label{fig-intro}
\end{figure*}

Out-of-distribution generalization in time series (TS-OOG) presents unique complexities compared to modalities like images or text, primarily because its distribution shifts are intrinsically tied to the temporal dimension \cite{222,khoee2024domain,lv2025large}. These shifts, driven by underlying temporal dynamics such as seasonality, long-term trends, systemic aging, or abrupt external shocks (e.g., policy changes, equipment failures) \cite{He2024TNNLS}, statistically manifest in two core ways. Fig.~\ref{fig-intro} visually illustrates how temporal dynamics lead to these distinct statistical shifts. The first type is covariate shift, where the distribution of input features, $\mathbb{P}(X)$, changes over time, but the relationship between features and the target variable, $\mathbb{P}(Y|X)$, remains stable \cite{Ott2022MM,RAY20229608}. The second, and often more challenging type, is concept drift. Here, the fundamental mapping between inputs and outputs, $\mathbb{P}(Y|X)$, itself changes \cite{cohen2008real,You2021CIKM,Sato2021,ijcai2024p888,Read2025}. For instance, a model trained to recognize a young athlete's gait patterns might fail when applied to an elderly person \cite{217}. This occurs because the underlying concept, specifically the relationship between sensor readings and gait, has fundamentally changed. Accurately identifying and adapting to these two forms of statistical distribution shift, both rooted in the evolving nature of time series data, is the central pursuit of TS-OOG research.

To address the challenges of distribution shifts in time series, the technical landscape of TS-OOG spans a spectrum from static to dynamic generalization approaches. Static generalization represents one end, aiming to train a fixed model during the training phase that remains effective under limited distribution shifts encountered during testing, without needing further adaptation. This passive robustness is often pursued through techniques like invariant representation learning, where models such as FOIL \cite{46} utilize invariant risk minimization (IRM) \cite{220} to identify stable patterns, and decoupling-based methods, exemplified by SCNN, which decompose time series into components presumed to have differing stability. In contrast, adaptive mechanisms tackle more dynamic environments by equipping models, like SOLID \cite{Chen2024KDD}, with the capability to actively adjust to ongoing temporal and concept drifts in real-time during deployment, rather than relying solely on the initial training. A distinct paradigm involves large time series models (LTSMs), like TimeGPT-1 \cite{79}. These leverage knowledge learned from massive, diverse datasets through large-scale pre-training, aiming to generalize to entirely new tasks or significant distribution shifts, often without task-specific fine-tuning.

\begin{figure*}
	\centering
	 \includegraphics[width=0.98\textwidth]{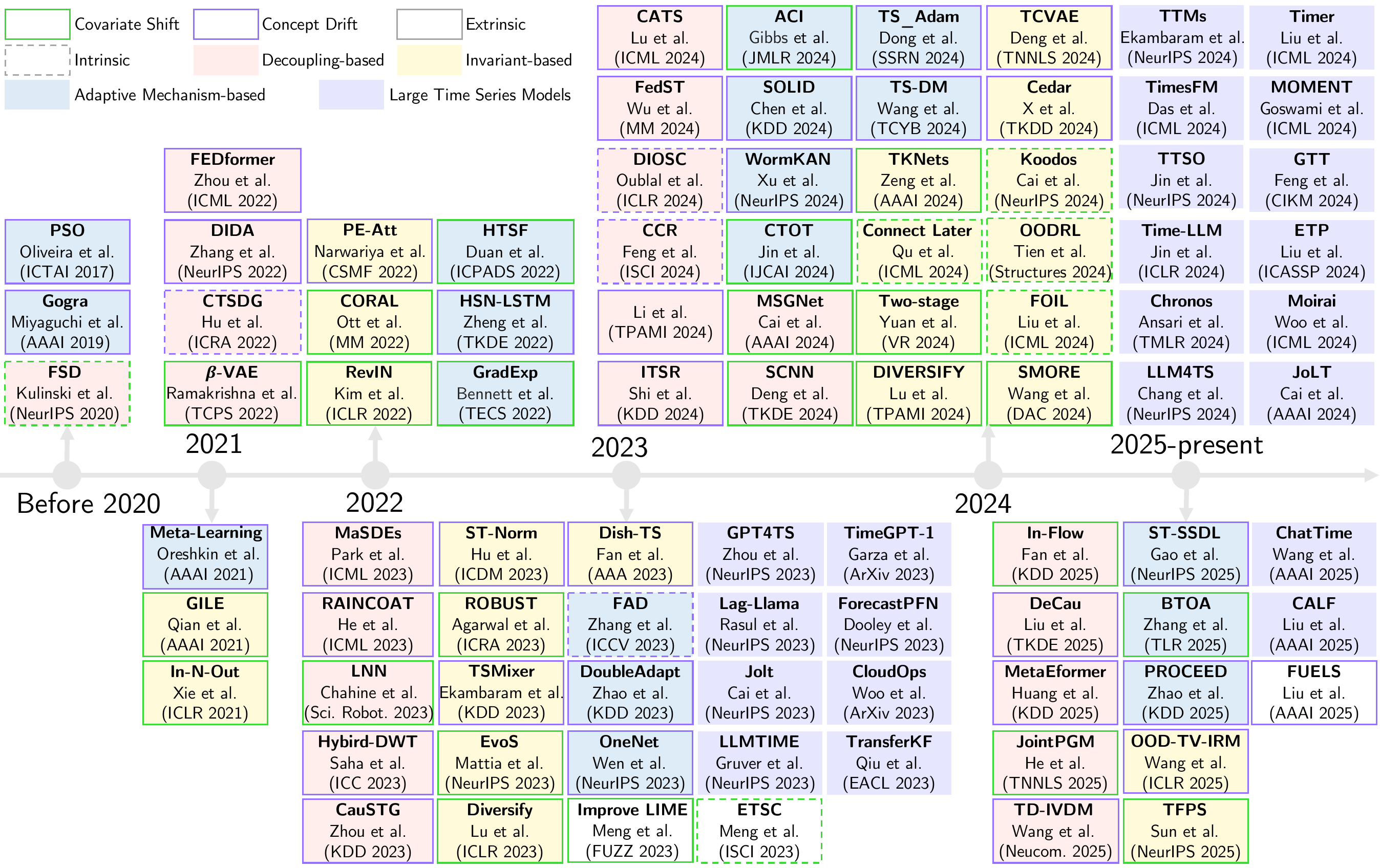}
	\caption{A research roadmap of representative methods for TS-OOG. The methods are categorized along three dimensions: data distribution assumptions (indicated by border color), representation learning strategies (indicated by line style), and OOD evaluation protocols (indicated by block color). Methods within the roadmap are presented in chronological order. For a more comprehensive summary, please refer to the main text.}
	\label{fig-time}
\end{figure*}

To navigate the growing complexity and importance of this field, this paper presents the comprehensive survey on TS-OOG, with a particular focus on data distributions, representation learning, and future research directions. To delineate the field's evolutionary trajectory, we organize representative TS-OOG methods in chronological order, as illustrated in Fig. \ref{fig-time}. Given that research methodologies have undergone continuous and rapid diversification since 2020, this chronological approach is essential for clearly tracing the field's development. By offering a structured synthesis of this dynamic landscape, our goal is to provide researchers with a clear roadmap of existing solutions, identify open challenges, and inspire further innovation in this critical area.

\textit{Related Surveys.} Our survey differs significantly from the existing literature in scope and technical focus. While several excellent surveys on OOD generalization exist, they do not specifically address the nuances of time series data. For instance, Liu et al. \cite{15} provide a broad overview of OOD generalization, emphasizing causal mechanisms and stable architectures for general data modalities. Concurrently, Li et al. \cite{17} focus on OOD challenges in graph-structured data, employing geometric learning paradigms. More recently, Yang et al. \cite{210} explored the OOD problem within the context of large-scale language models. In contrast, our work presents the systematic framework dedicated entirely to TS-OOG, a domain that poses a unique confluence of challenges, including non-stationarity, complex temporal dynamics, and time-varying feature interactions. By focusing exclusively on time series, we provide a more granular and domain-specific analysis than what is currently available.

\begin{figure}
	\centering
	 \includegraphics[width=0.42\textwidth]{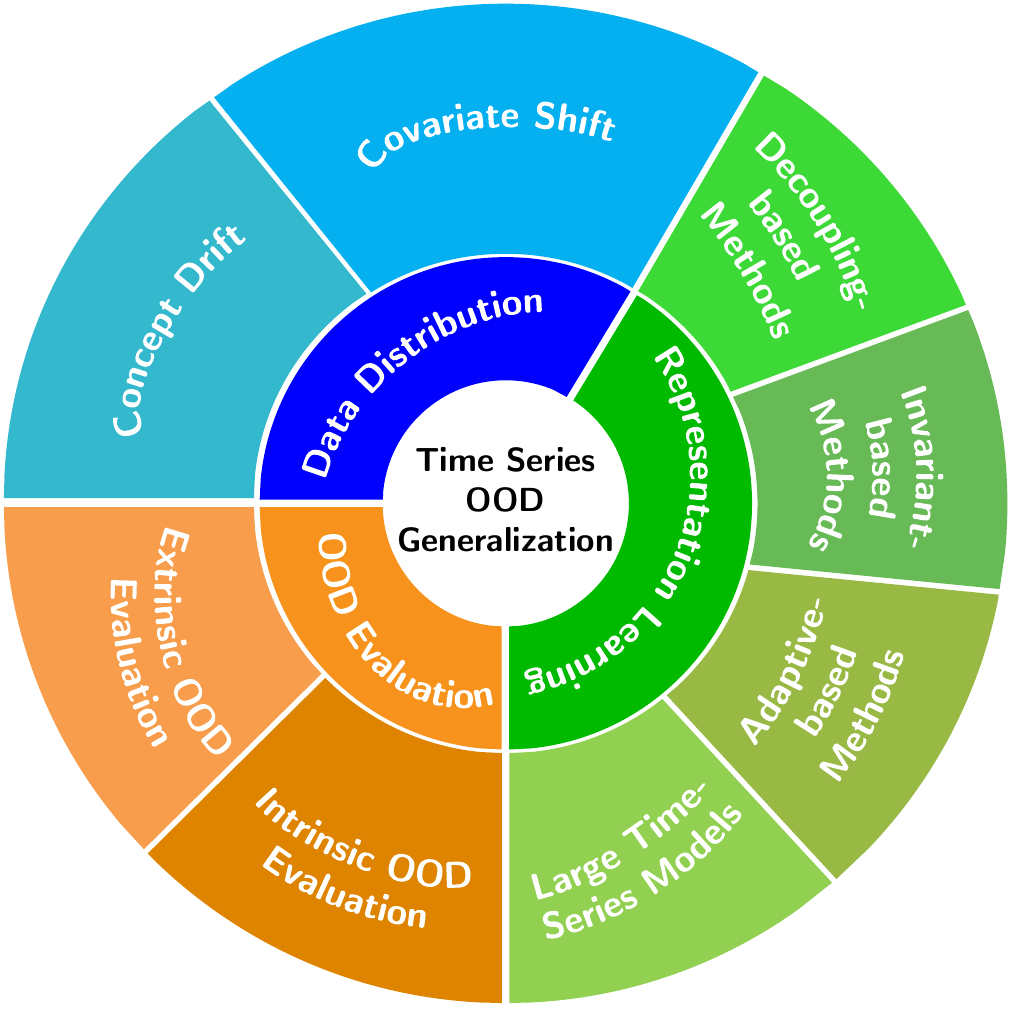}
	\caption{A comprehensive taxonomy of TS-OOG methods.}
	\label{fig-pie}
\end{figure}

\textit{Our Contributions.} This paper aims to fill the aforementioned void by providing the first comprehensive and systematic review of TS-OOG. Our primary contributions are as follows:
\begin{itemize}
    \item  We propose an innovative classification framework that organizes existing TS-OOG research along three key dimensions: data distribution, representation learning, and OOD evaluation, offering a clear lens through which to understand the field.
   \item We introduce and elaborate on several important and emerging categories, including decoupling-based methods, invariant-based methods, adaptive mechanism-based, and LTSMs, as illustrated in Fig. \ref{fig-pie}. Additionally, we incorporate a wide range of recent studies (over 55\% published in 2024 and 2025).
    \item We conduct a deep analysis of the foundations, technical details, and real-world applications of various approaches. Furthermore, we have established an open-source code repository\footnote[1]{\textcolor{blue}{\url{https://tsood-generalization.com}}} to facilitate future research and practical implementation in the community.
\end{itemize}

\textit{Organization.} The remainder of this paper is organized as follows: section  \ref{sec-pre} introduces the background of TS-OOG. Section \ref{sec-method} details our proposed taxonomy, the core components of which are elaborated upon in the subsequent chapters: section \ref{sec4_data} explores data distribution, section \ref{sec5_rep} is dedicated to representation learning, and section \ref{sec6_ood-evaluation} focuses on OOD evaluation. Following this detailed classification, sections \ref{sec7_applications} and  \ref{sec8_future} discuss application scenarios, challenges, and future directions, respectively. Finally, section \ref{sec9_con} concludes the survey.

%% file: 2_background.tex
\subsection{Definition of Time Series Data}
Time series is a sequence of data points ordered by time, typically representing measurements or observations taken at successive, equally spaced points in time. Mathematically, a time series \(X=\{X_t\}_{t=1}^{T}\) consists of \( T \) observations where each \( X_t \in \mathbb{R}^p \) represents the value of the time series at time \( t \), and \( p \) denotes the dimensionality of each observation (i.e., the number of features at each time step). Data augmentation operations such as the flipping operation generate a new sequence \( X_t' = -X_t \). For cases \cite{19,188,189}, the time series \( X_t \) is decomposed as \( X_t = \tau_t + s_t + r_t \), where \( \tau_t \) represents the trend component, \( s_t \) denotes the seasonal component, and \( r_t \) signifies the residual component.

Consider a time series training \( \mathcal{D} ^{\text{train}} = \{ (X_i, Y_i) \}_{i=1}^T \), where each \( X_i \in \mathcal{X} \subset \mathbb{R}^p \) is a \( p \)-dimensional instance, and \( Y_i \in Y = \{ 1, \cdots, C \} \) is its label. Assume the dataset comprises \( N \) latent domains, where \( 1 \leq N \leq T \). Each latent domain \( i \) has an associated joint distribution \( \mathbb{P} ^i(X, Y) \) over \( \mathcal{X} \times \mathcal{Y}\). The weights of these distributions are denoted by \( \pi_i \), satisfying \( \sum_{i=1}^N \pi_i = 1 \) and \( \pi_i > 0 \). The overall distribution of the time series is expressed as a weighted sum of the distributions across all latent domains:
\begin{equation}
\mathbb{P} ^{\text{train}}(X, Y) = \sum_{i=1}^{N} \pi_i \mathbb{P} ^i(X, Y).  \label{eq:mixture_dist}
\end{equation}
Here, \( \mathbb{P}^{\text{train}}(X, Y)\) aligns with the joint distribution of the dataset over the input-output space \( \mathcal{X} \times \mathcal{Y}\). Each latent domain is represented by a set of samples following its distribution. These samples can be freely interchanged within their respective domains.

We provide a list of commonly used notations in Table \ref{tab:notation}.

\subsection{Formalizing Generalization Challenges in Time Series}
To properly situate the contributions of various methodologies, it is crucial to formally define the distinct generalization challenges that arise in time series analysis. We delineate two primary problems, which are distinguished fundamentally by their underlying mathematical assumptions regarding the stability of the task's input and output spaces.

\subsubsection{OOD Generalization within a Fixed Time Series Context}
\label{sec:fixed_ts}
The classical formulation of OOD generalization addresses scenarios where a model must adapt to a new data distribution while the core task remains unchanged. This is mathematically defined by the assumption that the input space $\mathcal{X}$ and the output space $\mathcal{Y}$ are invariant between the training and test sets:
\begin{equation}
\mathcal{X}^{\text{train}} = \mathcal{X}^{\text{test}} \quad \text{and} \quad \mathcal{Y}^{\text{train}} = \mathcal{Y}^{\text{test}}. 
\label{eq:fixed_context_condition}
\end{equation}
The generalization challenge arises exclusively from a shift in the joint probability distribution of the data:
\begin{equation}
  \mathbb{P}^{\text{train}}(X,Y) \neq \mathbb{P}^{\text{test}}(X,Y). \label{eq:dist_shift} 
 \end{equation}

The uniqueness for time series lies in the nature of this distributional shift. Unlike static data such as images where a shift might be a change in lighting or background, the shift in time series is a manifestation of a dynamic, non-stationary data-generating process that evolves over time.

The objective is to train a model $f_{\theta}$ that minimizes the expected loss over future observations from this evolving process.
The general OOD objective,
$\min_{\theta} \allowbreak 
\mathbb{E}_{(X,Y)\sim\mathbb{P}_{\text{test}}} 
\allowbreak [\mathcal{L}(f_{\theta}(X),Y)]$, 
is therefore specialized for time series to explicitly model this forward-looking prediction task:
 \begin{equation}
 \min_{\theta} \mathbb{E}_{\{X_t, Y_t\} \sim \mathbb{P}_{\text{test}}} \left[ \sum_{t=T+1}^{T+\tau} \mathcal{L} (f_{\theta}(X_{t-\Delta:t-1}), Y_t) \right], \label{eq:ood_objective_fixed} 
 \end{equation}
where $\tau$ represents the prediction horizon (the number of future steps to forecast), and $\Delta$ denotes the look-back window (the number of past steps used as input). This formulation encapsulates the goal of adapting to temporal drifts for a single, well-defined task.

\begin{table}[h] 
\centering
\caption{Notations used in this survey.}
\label{tab:notation} 
\begin{tabularx}{\columnwidth}{p{0.9cm}X|p{0.9cm}X}
\toprule
Notation & Description & Notation & Description \\
\midrule
\( X_t \) & Observations at time point \( t \) &  \( y \) & Label \\ 
\( Y_t \) & Predictions at time point \( t \)& \( \mathcal{Y} \) & Label space\\
\( \mathcal{X} \)  & Input space &  \(  \mathcal{L}  \) & Loss function\\
\( \pi_i \) & The weight of the \( i \)-th domain & \( \theta \) & Parameter \\
\( \mathcal{D} \)  &Domain & \( f(\cdot) \) & Predictive model\\ 
\( \mathbb{P} (\cdot) \) & Distribution &$ \mathbb{E}\left ( \cdot\right ) $ & Expectation \\
\bottomrule
\end{tabularx}
\end{table}

\subsubsection{Cross-Domain and Cross-Task Generalization} 
\label{sec:cross_task_gen}

A more profound generalization challenge arises when models must operate across different domains or tasks where the core assumption of fixed input and output spaces is relinquished. This type of generalization, extensively studied in the fields of transfer learning and meta-learning, requires models to transfer knowledge learned from potentially diverse source datasets ($S_i$) to new target tasks ($\text{test}$) that may be structurally different.

Mathematically, this scenario is characterized by a departure from the assumption in Eq.~\eqref{eq:fixed_context_condition}. The input spaces ($\mathcal{X}$) and/or output spaces ($\mathcal{Y}$) may differ between source domains/tasks ($S_i$) and the target domain/task ($\text{test}$): 
\begin{equation}
\mathcal{X}^{S_i} \neq \mathcal{X}^{\text{test}} \quad \text{and/or} \quad \mathcal{Y}^{S_i} \neq \mathcal{Y}^{\text{test}}. \label{eq:cross_task_condition}
\end{equation}

In the context of time series, this mathematical inequality has profound physical meaning. A difference in input spaces ($\mathcal{X}^{S_i} \neq \mathcal{X}^{\text{test}}$) means the model must generalize across time series with different numbers of variates, different physical units, and entirely different feature sets, for instance from financial market data to clinical ECG signals. A difference in output spaces ($\mathcal{Y}^{S_i} \neq \mathcal{X}^{\text{test}}$) means the task itself changes, for example from a regression problem like forecasting to a classification problem like anomaly detection.

The objective in this setting extends beyond mere robustness to distributional shift within a single task (as described in Sec. \ref{sec:fixed_ts}). Instead, the primary goal is typically to learn universal representations of temporal dynamics (e.g., trends, seasonality, anomalies, causal relations) during a large-scale pre-training phase on diverse source data. These representations should be sufficiently abstract and informative to be effectively redeployed (often via zero-shot inference or few-shot fine-tuning) across disparate target domains and tasks defined by Eq.~\eqref{eq:cross_task_condition}. 

Unlike the fixed-context OOD problem which has a clear objective function like Eq.~\eqref{eq:ood_objective_fixed} focused on minimizing loss for a specific future prediction task, a single, equivalent objective function for the general cross-domain/cross-task setting is less straightforward to formulate, especially for zero-shot generalization. The objective is often implicitly defined by the pre-training strategy (e.g., masked prediction loss across diverse series, contrastive objectives) which aims to produce representations that exhibit strong transferability. Performance is typically measured post-hoc via evaluation protocols on downstream target tasks. Consequently, the objective function defined in Eq.~\eqref{eq:ood_objective_fixed} is generally not applicable to this scenario due to the potential changes in input/output spaces and task definition.

While methods like decoupling and invariant learning primarily target the challenge outlined in Sec. \ref{sec:fixed_ts}, LTSMs, leveraging extensive pre-training, have emerged as a prominent approach particularly aimed at tackling this more ambitious cross-domain and cross-task generalization problem, as will be discussed further in Sec. \ref{sec5_ltsm}. It is worth noting, however, that LTSMs can also be applied to the fixed-context OOD problem (Sec.~\ref{sec:fixed_ts}), where their pre-trained representations might offer enhanced robustness against distributional shifts compared to models trained solely on the target task's historical data.

%% file: fig_tree.tex
\begin{forest}
  for tree={
  grow=east,
  reversed=true,
  anchor=base west,
  parent anchor=east,
  child anchor=west,
  base=left,
  font=\small,
  rectangle,
  draw,
  rounded corners,align=left,
  minimum width=2.5em,
  inner xsep=4pt,
  inner ysep=1pt,
  },
  where level=1{text width=5.3em,fill=blue!10}{},
  where level=2{text width=5em,font=\footnotesize,fill=pink!30}{},
  where level=3{font=\footnotesize,yshift=0.26pt,fill=yellow!20}{},
  [Out-of-distribution \\ Generalization \\ in Time Series,fill=green!20
        [Data Distribution\\(Sec. \ref{sec4_data}),text width=7em
            [Covariate Shift, text width=5.4em [ BTOA \cite{zhang2025batch} / Improved LIME \cite{22} / \cite{Ott2022MM} /SEGAL \cite{23}/   PE-Att \cite{87} /\\ Two-stage method \cite{24} /  TKNets \cite{25} / CTOT \cite{26} / RevIN \cite{27} /  \\ Connect Later \cite{28} / ETSC \cite{29} / TIMEX++ \cite{30} / TFPS \cite{sun2025learningpatternspecificexpertstime} / \\  IN-Flow \cite{Fan2025kdd} / HTSF \cite{Duan2023} /  / GradExp\cite{Stefanos2022ICML} / JointPGM \cite{He2025TNNLS} /  FSD \cite{36}. 
            	]
            ]
            [Concept Drift, text width=5.2em
            [  DIOSC \cite{33} / RAINCOAT \cite{216} / MetaEformer \cite{Huang2025MetaEformer} / TCVAE \cite{He2024TNNLS} / \cite{read2018concept} /   INNT \cite{dengalleviating} /  \\SOLID \cite{Chen2024KDD} / OneNet \cite{Zhang2023OneNet} / CATS \cite{Lu2024CATS} / DIDA \cite{Zhang2022NeurIPS} / \cite{masserano2022adaptive} / TS\_Adam \cite{dong5570331rethinking} /   \cite{Li2024TPAMI} /  \\ DeCau \cite{Liu2025tkde} /   WormKAN \cite{xu2024kandrift} / PROCEED \cite{zhao2025kdd} / Cogra  \cite{Miyaguchi2019aaai} / HSN-LSTM \cite{Zheng2022tkde} /  \\ CORAL \cite{xu2025coralconceptdriftrepresentation} / TD-IVDM \cite{WANG2025131120} / \cite{Forman2006} / \cite{oliveira2025dynamic} / TS-DM \cite{Wang2024TCYB} /  CatSight \cite{florez2023catsight} / PSO \cite{Oliveira2017} / \\ Autoencoder-ADWIN-LSTM \cite{Saravana2025} / OS-ELM \cite{Yang2020tnnls} / ST-SSDL \cite{gao2025different} / CEP \cite{zhan2025continuousevolutionpooltaming} . 
             ]
            ]
        ]
        [Representation\\ Learning (Sec. \ref{sec5_rep}),text width=7em
          [Decoupling-based,text width=6.4em
              [\emph{Multi-Structured Analysis:}  ITSR \cite{34} / SCNN \cite{35} /   FSD \cite{36} /   \(\beta\)-VAE \cite{38} /   \\ MSGNet \cite{40} / JointPGM \cite{He2025TNNLS} / Hybird-DWT  \cite{60}/MetaEformer \cite{Huang2025MetaEformer} / FEDformer \cite{zhou2022fedformer}. 
              ] 
              [\emph{Causality-Inspired:}  CTSDG \cite{42} /   MaSDEs \cite{43} /  LNN \cite{44} / \\ CCR \cite{45} / DIOSC \cite{33} / \cite{Li2024TPAMI} / DIDA \cite{Zhang2022NeurIPS} / DeCau \cite{Liu2025tkde}/CauSTG \cite{64}.
              ]
          ]
          [Invariant-based, text width=5.4em
            [\emph{Invariant Risk Minimization:} FOIL \cite{46} / ROBUST \cite{47} /  TSModel  \cite{48} /  \\ TSMixer  \cite{50} /  OOD-TV-IRM\cite{207} / ST-Norm \cite{Hu2023icdm} / Cedar \cite{Deng2024} / TFPS \cite{sun2025learningpatternspecificexpertstime}.
            ]
            [\emph{Domain-Invariance:} Two-stage method \cite{24} / Connect Later \cite{28} /  \\ TKNets \cite{25}  / RAINCOAT \cite{216} / Koodos \cite{52} /  Diversify  \cite{53,54} /  \\ OODRL \cite{55} /  In-N-Out \cite{56} / SMORE \cite{57} / EvoS \cite{58} / RevIN \cite{27} /\\ GILE \cite{213} / \cite{Ott2022MM} / TCVAE \cite{He2024TNNLS} / INNT \cite{dengalleviating} / Dish-TS \cite{Fan2023aaai}  /  PE-Att \cite{87}.
            ]
          ]
           [Adaptive \\ Mechanism-based, text width=6.7em
                          [\emph{Explicit:} \cite{Oliveira2017} / \cite{Isaac2024} / SOLID \cite{Chen2024KDD} / BTOA \cite{zhang2025batch} / HTSF \cite{Duan2023} / \\ DoubleAdapt  \cite{49} / CEP \cite{zhan2025continuousevolutionpooltaming} /  PROCEED \cite{zhao2025kdd} / ST-SSDL \cite{gao2025different} /\\GradExp \cite{Stefanos2022ICML} / CTOT \cite{26} / \cite{read2018concept} / \cite{oliveira2025dynamic} / PSO \cite{Oliveira2017} / OneNet \cite{Zhang2023OneNet}.
            ]
                        [\emph{Implicit:} TS\_Adam \cite{dong5570331rethinking} / Cogra \cite{Miyaguchi2019aaai} / WormKAN \cite{xu2024kandrift} / \\ CORAL \cite{xu2025coralconceptdriftrepresentation}  / HSN-LSTM \cite{Zheng2022tkde} / Meta-Learning  \cite{51} / FAD \cite{89}.
            ]
            ]
              [Large Time \\ Series Models, text width=5.2em
                          [\emph{LLM-based:} (Tuning-based) Time-LLM \cite{70} / Moirai \cite{71} /  ChatTime \cite{76} / \\ GPT4TS \cite{81} / CALF \cite{75}  /  TransferKF \cite{68} / ETP \cite{69} / LLM4TS \cite{72}; \\ 
                          (Prompt-based)  LLMTIME \cite{77} / CloudOps \cite{78} / TableTime \cite{80}.
            ]
                        [\emph{Time Series Foundation Models:}  PatchTST \cite{67} / ForecastPFN \cite{65} \\ Jolt \cite{66} / GTT \cite{144} / TimeGPT-1 \cite{79} /  MOMENT \cite{83} / Timer \cite{84} / \\ TimesFM \cite{85} / Chronos \cite{86} /  TTMs \cite{73} / Lag-Llama \cite{82} / TimeRAF \cite{11031238}.
            ]
          ]
        ]
        [OOD Evaluation\\(Sec. \ref{sec6_ood-evaluation}),text width=7em
          [Extrinsic, text width=3em
            [\emph{Verifiable:} Improved LIME \cite{22} /  SEGAL \cite{23} /  TIMEX++ \cite{30}  /   CauSTG \cite{64} / SCNN \cite{35}/  \\ MSGNet \cite{40} /  MetaEformer \cite{Huang2025MetaEformer} / FEDformer \cite{zhou2022fedformer} / \cite{Li2024TPAMI}/ DIDA \cite{Zhang2022NeurIPS} /  DeCau \cite{Liu2025tkde} /   \cite{Ott2022MM} / \\ TCVAE \cite{He2024TNNLS} / INNT \cite{dengalleviating} / ST-Norm \cite{Hu2023icdm} / Cedar \cite{Deng2024} / TFPS \cite{sun2025learningpatternspecificexpertstime} /JointPGM \cite{He2025TNNLS}  / \cite{masserano2022adaptive} /  \\ Cogra  \cite{Miyaguchi2019aaai} / HTSF \cite{Duan2023} / BTOA \cite{zhang2025batch} / GradExp \cite{Stefanos2022ICML} /\cite{read2018concept} /  OneNet \cite{Zhang2023OneNet} /CEP \cite{zhan2025continuousevolutionpooltaming}/ \\PSO \cite{Oliveira2017} / \cite{Forman2006} / \cite{oliveira2025dynamic} / TS-DM \cite{Wang2024TCYB} /  ST-SSDL \cite{gao2025different}/  HSN-LSTM \cite{Zheng2022tkde} / CORAL \cite{xu2025coralconceptdriftrepresentation} / \\  TS\_Adam \cite{dong5570331rethinking} / WormKAN \cite{xu2024kandrift} / FUELS \cite{Liu2025FUELS} / Dish-TS \cite{Fan2023aaai} / PROCEED \cite{zhao2025kdd} /RevIN \cite{27} .
             ]
            [\emph{Speculative:} In-N-Out \cite{56} / SMORE \cite{57} /  PE-Att \cite{87} / \\  Two-stage method \cite{24}  /  LNN \cite{44} /  VRStrokeOOD\cite{24}.
             ]
           ]
          [Intrinsic, text width=3em
            [ OODRL \cite{55} / DIVERSIFY \cite{53,54} /   CCR \cite{45} /    Koodos \cite{52} /    FAD \cite{89} / FOIL  \cite{46} /  \\ CTSDG  \cite{42}  /  Connect  Later  \cite{28} /   EvoS \cite{58}  / DIOSC  \cite{33}/ TD-IVDM \cite{WANG2025131120} /  \\ CatSight \cite{florez2023catsight} / Autoencoder-ADWIN-LSTM \cite{Saravana2025} / \cite{Yang2020tnnls} /OS-ELM \cite{Yang2020tnnls} / FSD \cite{36}. 
            ]
           ]
       ]
    ]
 ]
\end{forest}

%% file: 4_sec_data_distribution.tex
\begin{figure}
	\centering
	 \includegraphics[width=0.48\textwidth]{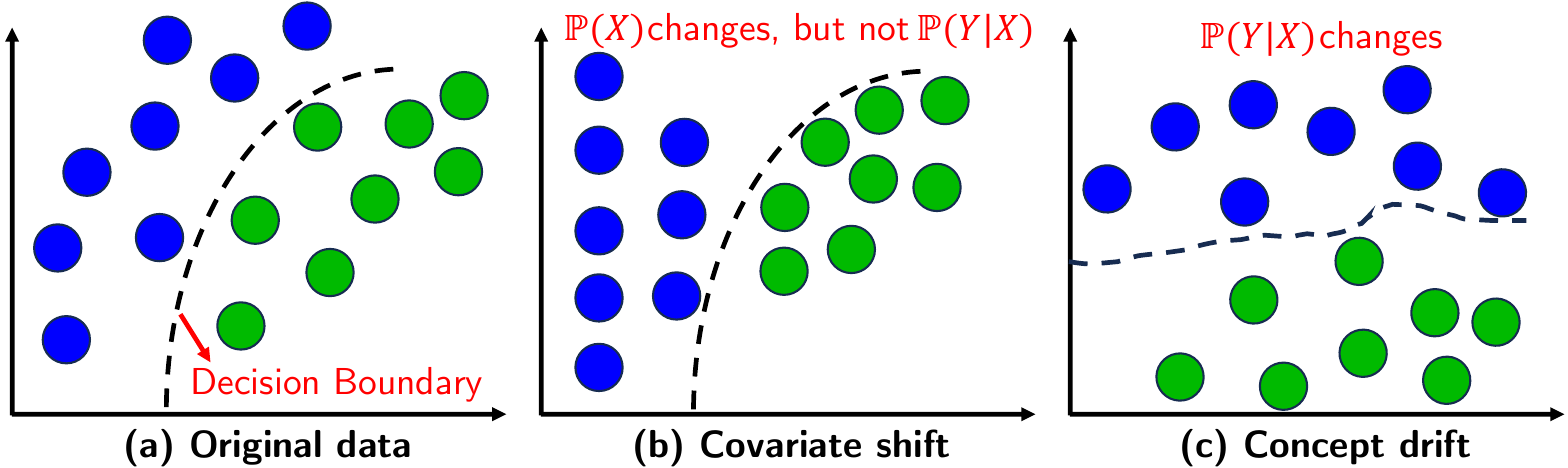}
	\caption{Each circle represents a data instance, with different colors indicating the class to which the instance belongs. \textbf{(a). Original data:} The distribution of instances and the classification boundary (dashed line) remain stable. \textbf{(b). Covariate shift:} The distribution of instances changes (the feature distribution \( \mathbb{P} (X) \) changes), but the class decision boundary \( \mathbb{P} (Y|X) \) remains unchanged. \textbf{(c). Concept drift:} The class decision boundary \( \mathbb{P} (Y|X) \) changes, but the distribution of instances (the feature distribution \( \mathbb{P} (X) \)) remains unchanged.}
	\label{fig-data_distribution}
\end{figure}

The distribution of data is a critical factor that determines a model's generalization ability and robustness. However, real-world data is often dynamic and evolving rather than static, leading to distribution shifts across different stages, scenarios, or environments. Distribution shift violates the independent and identically distributed (i.i.d.) assumption in machine learning, posing significant challenges to model performance. The primary types of distribution shift include covariate shift and concept drift, as illustrated in Fig. \ref{fig-data_distribution}.

\subsection{Covariate Shift}
Covariate shift refers to a change in the input feature distribution \( \mathbb{P} (X) \) while the conditional distribution \( \mathbb{P} (Y|X) \) remains constant. Its fundamental impact is that the feature relationships learned by a model on the training set may no longer hold in the testing environment, thereby impairing generalization performance. To address this, the research community has proposed various strategies from the perspectives of distribution modeling, feature alignment, and representation learning.

On one hand, generative modeling and data augmentation methods enhance model robustness by expanding the training distribution and increasing sample diversity. For instance, conditional variational autoencoders \cite{He2025TNNLS} further mitigate distribution shifts caused by feature replacement by generating samples consistent with the original distribution \cite{21}.

On the other hand, with the rise of time series and cross-domain tasks, research on covariate shift has increasingly emphasized dynamic and domain-adaptive solutions. Typical approaches include domain adaptation methods based on optimal transport or correlation alignment to learn domain-invariant embeddings, and input distribution stabilization strategies based on normalization and transformation mechanisms, such as removing non-stationary components from time series to maintain predictive stability \cite{27}. Furthermore, recently proposed frameworks featuring decoupled and adaptive mechanisms (e.g., \cite{sun2025learningpatternspecificexpertstime, He2025TNNLS,Duan2023,zhang2025batch,Stefanos2022ICML}).

\subsection{Concept Drift}
Unlike covariate shift, concept drift refers to scenarios where the relationship between inputs and outputs, \( \mathbb{P} (Y|X) \), changes \cite{31}. This change typically stems from an evolution in the underlying data generation process or system semantics. For example, in time series tasks, a change in the correlation between attributes can directly lead to a decline in model performance in subsequent phases \cite{33}. The fundamental challenge of concept drift is that it involves not just a change in input features, but a transformation of the target patterns themselves.

Research on addressing concept drift primarily focuses on three main ideas: structural decoupling, feature invariance learning, and adaptive mechanisms. First, decoupled modeling aims to capture stable patterns by separating modes at different semantic levels. For example, MetaEformer decomposes time series into trend and seasonal components to extract meta-patterns \cite{Huang2025MetaEformer}, while CATS \cite{Lu2024CATS} forms new exogenous variables by decoupling inter-sequence correlations \cite{florez2023catsight}. Additionally, works from a causal perspective (e.g., DIDA \cite{Zhang2022NeurIPS}, DeCau \cite{Liu2025tkde}, and transferable time series forecasting under causal conditional shift \cite{Li2024TPAMI}) improve generalization by distinguishing between causal and spurious correlations. Second, invariant modeling seeks to learn latent representations that are stable across different distributions. For example, TCVAE \cite{He2024TNNLS} and INNT \cite{dengalleviating} achieve temporal distribution smoothing through latent space mapping. Finally, adaptive and drift detection mechanisms provide feasible solutions for concept drift in dynamic environments. Continuous gradient updates \cite{read2018concept}, real-time calibration at the inference stage \cite{Chen2024KDD, zhao2025kdd}, and optimization-based parameter tuning \cite{Oliveira2017} have all been proven effective. Unsupervised detection methods, such as CatSight \cite{florez2023catsight} and Autoencoder-ADWIN-LSTM \cite{Saravana2025}, monitor potential concept changes by analyzing prediction or reconstruction errors, thereby establishing a prerequisite for model adaptation.

In many real-world scenarios, covariate and concept drifts often coexist and are intertwined. Consequently, research is gradually shifting towards designing unified and robust modeling frameworks to simultaneously address distributional changes at both the input and semantic levels. Typical works such as Dish-TS \cite{Fan2023aaai} and ST-Norm \cite{Hu2023icdm} use spatio-temporal normalization to reconcile spatial heterogeneity and temporal dynamics. FEDformer \cite{Hu2023icdm} stabilizes statistical properties while capturing long-term dependencies. Meanwhile, Cedar \cite{Deng2024} and ACI \cite{Gibbs2024} approach the problem from a domain generalization perspective, building general-purpose models adaptable to hybrid shifts. This trend signifies that the study of distribution learning is advancing towards a more holistic and dynamic paradigm.

%% file: 5_sec_representation_learning.tex
In time series data, distribution shifts due to seasonality, trends, or unexpected events often cause mismatches between training and test data. Representation learning is crucial for TS-OOG as it extracts robust features to enhance generalization. Tables \ref{tab:tsmethods} and \ref{tab:models_summary} summarize the key characteristics of different approaches.

\subsection{Decoupling-based Methods} Decoupling-based methods address distribution shifts by separating time series into relevant and irrelevant features, thereby improving model robustness and generalization across different temporal environments \cite{Wu2024kdd,Lu2024CATS,Fan2025kdd}. This ensures performance stability. Two major strategies are introduced below: multi-structured analysis and causality-inspired approaches.

\input{table_methods}

\subsubsection{Multi-Structured Analysis} 

Multi-structured analysis methods based on decoupling decompose time series data (e.g., through label decomposition or component decomposition) into multiple independent and interpretable parts such as short-term, long-term, and seasonal components. For example, ITSR \cite{34} applies a learnable orthogonal decomposition to partition time series into invariant and correlated features, ensuring consistency of core patterns across different domains while mitigating domain shifts. SCNN \cite{35} decomposes multivariate time series into structured and heterogeneous components, with dedicated sub-networks modeling each component to predict complex spatiotemporal patterns efficiently. In addition, FSD \cite{36} leverages conditional distribution tests to locate features where shifts occur, thus enhancing robustness in high-dimensional and noisy environments. \(\beta\)-VAE \cite{38} uses latent-space modeling for efficient OOD detection, sensitively capturing changes in image features through partial decomposition of latent variables, thereby identifying OOD samples. JointPGM \cite{He2025TNNLS} decomposes the time-variant distributional shift into intraseries and interseries components, which are then modeled by distinct learners. MetaEformer \cite{Huang2025MetaEformer} utilizes seasonal-trend decomposition to extract and purify fundamental waveforms known as meta-patterns, which are then adaptively leveraged to reconstruct future load patterns. FEDformer \cite{zhou2022fedformer}  combines a seasonal-trend decomposition method to capture the global profile of a time series with a frequency-enhanced Transformer designed to model more detailed structures.  Finally, MSGNet \cite{40} employs frequency-domain analysis to decompose time series into multi-scale components and combines adaptive graph convolutions to capture cross-series correlations at different temporal scales, thus boosting both the accuracy of multivariate time series forecasting and generalization to OOD samples.

\subsubsection{Causality-Inspired Models} 
The decoupling-based causal approaches mitigate spurious correlations in time series by isolating and extracting causal features, thereby significantly enhancing the model's generalization performance on unseen distributions. Progressively moving from raw causal structures to extracting causal features subject to less local distribution shift, then to extracting causal features under more substantial global distribution shifts, these methods gradually enhance adaptability to complex time series \cite{Li2024TPAMI}. For instance, Causal-HMM \cite{41} divides hidden variables into disease-related and spurious components, establishing a foundation for robust prediction by isolating causal features. Building on this, CTSDG \cite{42} introduces structural causal models combined with cyclic latent variable models, further strengthening the modeling of causal relationships in complex time series. To handle uncertainty and noise in time series, MaSDEs \cite{43} adopt a neural stochastic differential game framework, employing multi-agent collaboration to achieve more robust causal feature extraction. LNN \cite{44} takes advantage of liquid neural networks' dynamic causal modeling capabilities, improving adaptability and robustness in dynamic environments. As methods evolve, CCR \cite{45} focuses on extracting causal features related to external disturbances through causal representation learning, enhancing predictive power under exogenous conditions. Similarly, DIDA \cite{Zhang2022NeurIPS} disentangles spatio-temporal data into invariant and variant patterns, using a causal intervention mechanism to eliminate the spurious impacts of variant patterns and enhance generalization. DeCau \cite{Liu2025tkde} decomposes urban flow data into seasonal and trend components, attributing distribution shifts in each part to distinct context and structural factors, and then applies causal intervention and adjustment accordingly.  Finally, DIOSC \cite{33} incorporates contrastive learning to optimize the causal representation of time series under independence constraints, ensuring strong model generalization across diverse scenarios.

\subsection{Invariant-based Methods} Invariant-based methods aim to extract feature representations that remain stable across different environments in dynamic time series data, thereby addressing the challenges posed by distribution shifts. These approaches typically follow two complementary paradigms: one constrains the model to rely on invariant features via an optimized objective function (e.g., IRM), while the other eliminates domain discrepancies in the feature space (e.g., domain-invariance learning).

\subsubsection{Invariant Risk Minimization} 
The core idea of IRM is to discover feature representations that remain invariant across all training environments, thereby reducing the model’s reliance on spurious correlations specific to particular environments. For instance, FOIL \cite{46} proposes a model-agnostic solution for invariant learning in time series forecasting through a surrogate loss function and an environment inference module. A game-theoretic robust forecasting method \cite{47} applies the IRM concept by simulating adversarial perturbations to historical data. Moreover, Zhang et al. \cite{48} combine a specialized model architecture with IRM to capture cross-domain invariant features, further boosting generalization performance in time series. Cedar \cite{Deng2024} proposes a domain discrepancy regularization term to enforce consistent forecasting performance across different domains, thereby improving generalization. TFPS \cite{sun2025learningpatternspecificexpertstime} tackles patch-level distribution shifts by learning multiple stable predictive functions, each specialized for a distinct data pattern, thus aligning with the IRM goal of finding invariant predictors. Wang et al. \cite{207} leverage the total variation model and the primal-dual optimization method to enhance the model's generalization ability in OOD scenarios. By learning consistent features across diverse environments, IRM helps models maintain robust performance when facing unseen periods or sudden events. In time series applications, combining IRM methods with temporal structure modeling, dynamic invariant feature learning, meta-learning, adversarial training, and generative models have become a powerful toolkit for improving forecasting accuracy and generalization \cite{49,50,51}.

\subsubsection{Domain-Invariance Models}  
Domain-invariance methods seek to construct universal feature representations that enable models to maintain stable performance across different domains or environments, thereby mitigating performance degradation caused by distribution shifts  \cite{Ott2022MM}. First, they optimize the feature representations to maintain consistency across domains, effectively tackling distribution shifts \cite{24,28}. For example, conditional generative adversarial networks can discretize the output space and learn conditionally domain-invariant representations, improving adaptation to different interactive environments \cite{24}. Furthermore, techniques such as the Koopman operator can map complex nonlinear temporal dynamics to linear spaces, bolstering model adaptability and stability in changing environments \cite{25}. For example, Koodos \cite{52} incorporates the Koopman operator to linearize time evolution, combined with a multi-scale attention mechanism to capture feature distribution evolution. This approach enables models to remain stable and adaptable over continuous temporal changes, effectively handling distribution shifts in dynamic environments. Likewise, ST-Norm \cite{Hu2023icdm} uses invertible normalizing flows to transform spatio-temporal data into a unified, stationary distribution, thereby removing both spatial and temporal shifts. TCVAE \cite{He2024TNNLS} employs a temporal conditional variational autoencoder to model dynamic distributional dependencies, adapting to distributional shift by conditioning latent variables on temporal factors. INNT \cite{dengalleviating} also utilizes an invertible neural network to map time series into a latent space, where a pre-training strategy explicitly minimizes the distribution divergence between historical and future data. Additionally, some methods maximize distributional diversity to preserve heterogeneity while minimizing differences to reinforce domain invariance \cite{53}. Lu et al. \cite{54} propose an iterative framework that identifies worst-case potential distribution scenarios and reduces discrepancies. Furthermore, targeted enhancement strategies and adversarial training introduced during the fine-tuning phase improve the robustness of the model and the performance of the cross-domain, ensuring efficiency on unseen distributions \cite{28,55,56}. Applying hyperdimensional computing and multi-scale attention mechanisms enhances generalization when dealing with multi-sensor data and continuous distribution shifts \cite{57,58}.

\subsection{Adaptive Mechanism-based Methods}
While the aforementioned categories of representation learning methods focus on enhancing a model's static generalization capabilities, this section delves into strategies for adaptive representation learning in dynamic environments. As time series distributions continuously evolve, relying solely on static representations acquired during training is often insufficient to maintain stable performance.

Representation learning methods based on adaptive mechanisms are centered on a model's capacity for dynamic response and continuous adjustment in the face of distributional shifts in time series data. The core principle is to embed adjustable adaptive mechanisms within the model's architecture, optimization process, or inference phase. This enables the model to proactively update itself in response to evolving environmental and data distributions, thus ensuring stable performance and consistent representations in non-stationary environments. Broadly, these methods can be categorized into two primary approaches: explicit adaptation mechanisms and implicit adaptation mechanisms.

\subsubsection{Explicit Adaptation Mechanisms} 

Explicit adaptation mechanisms operate directly on a model's parameters or output process, enabling a rapid response to distributional shifts via explicit update strategies. These methods typically leverage incoming data streams during the testing or deployment phase for real-time model adjustments to maintain alignment with the current environment \cite{Oliveira2017,Isaac2024}. For example, SOLID \cite{Chen2024KDD} dynamically constructs contextually similar samples to achieve local distribution alignment, BTOA \cite{zhang2025batch} employs online data augmentation to mitigate concept drift, and PROCEED \cite{zhao2025kdd} proactively predicts and compensates for potential distribution shifts to perform lightweight model calibration during inference. Similarly, CEP \cite{zhan2025continuousevolutionpooltaming} tackles recurring concept drift by maintaining a dynamic pool of specialized forecasters, retrieving an existing forecaster for known patterns, and evolving a new one when a new concept is detected, thereby retaining knowledge of past concepts. Furthermore, ST-SSDL \cite{gao2025different}, through self-supervised deviation modeling, combines contrastive and deviation losses to enable the model to quantify the dynamic discrepancy between current inputs and historical patterns. This allows it to identify inputs that deviate from these patterns during inference, thereby facilitating rapid adaptation to non-stationary environments.

In addition to inference-time adjustments, some research embeds adaptability more deeply into the training process, giving rise to dynamic model and meta-learning frameworks \cite{Forman2006,read2018concept,masserano2022adaptive,Wang2024TCYB}. Such approaches learn not only the primary task but also the strategy of how to adapt to changes in that task \cite{oliveira2025dynamic}. For instance, the HTSF \cite{Duan2023} framework uses a hypernetwork to dynamically generate the main network's parameters based on the input data distribution, allowing the model architecture to reconfigure itself automatically in different environments. Methods like GradExp \cite{Stefanos2022ICML} employ mechanisms such as continuous adaptation, differentiable forgetting, and adaptive sampling to dynamically balance the weights of new and old knowledge, enabling the model to maintain its learning plasticity and generalization performance in long-term, evolving non-stationary environments.

\subsubsection{Implicit Adaptation Mechanisms} 

Implicit adaptation mechanisms, in contrast, embed intrinsic plasticity within the model's underlying architecture or optimization process. This enables the model to self-adjust in response to environmental changes without the need for explicit external intervention. These approaches are typically realized through innovations in optimizer design or representational structure, rendering adaptability an intrinsic property of the model rather than an externally applied operation.

At the optimization level, methods like TS\_Adam \cite{dong5570331rethinking} and Cogra \cite{Miyaguchi2019aaai} enhance the stability and responsiveness of the optimization process. TS\_Adam achieves this by improving the bias correction mechanism of the Adam optimizer, while Cogra introduces a novel sequential mean tracker, both allowing parameter updates to respond more sensitively to the dynamic changes of a non-stationary loss landscape. At the representation level, WormKAN \cite{xu2024kandrift} and CORAL \cite{xu2025coralconceptdriftrepresentation} learn the geometric structure of the latent space, such that concept drift manifests as analytically tractable and traceable changes within the internal representation. Meanwhile, HSN-LSTM \cite{Zheng2022tkde} integrates the computational paradigms of spiking and recurrent neurons, endowing the model with high sensitivity to dynamic inputs and structural adaptability at the network architecture level.

Compared to explicit mechanisms, implicit approaches place a greater emphasis on self-organization and gradual evolution, equipping the model with more enduring adaptive potential in long-term dynamic environments.

\subsection{Large Time Series Models}
\label{sec5_ltsm}

\begin{figure}
	\centering
	 \includegraphics[width=0.48\textwidth]{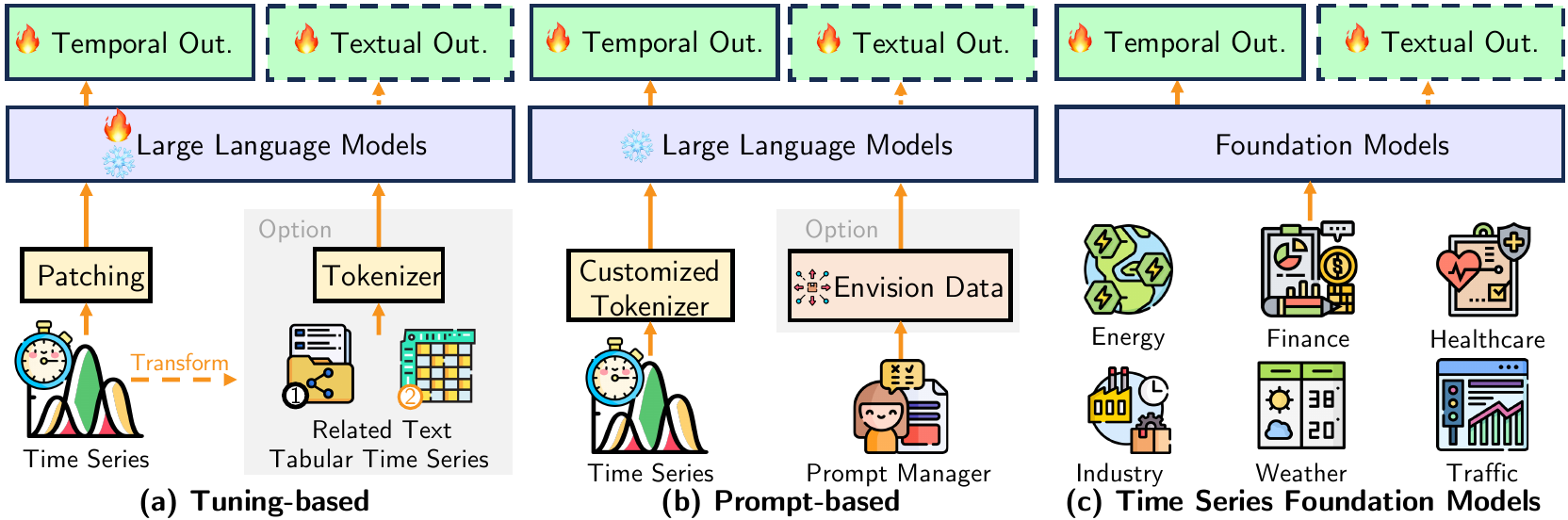}
	\caption{A taxonomy of LTSMs. \textbf{LLM-based methods:} These methods leverage pre-trained LLMs. Time series data is first converted into a token format that LLMs can understand, via techniques like patching (for tuning-based methods) or textual prompts (for prompt-based methods). The LLM is then used to perform the time series task without being retrained from scratch.  \textbf{Time series foundation models:} These are foundation models specifically designed for and pre-trained from the ground up on large-scale, diverse time series corpora from various domains (e.g., energy, finance, healthcare). They learn universal temporal representations directly from numerical data.}
	\label{fig-llm}
\end{figure}

LTSMs constitute a paradigm designed to tackle the broader challenge of cross-task and cross-domain generalization, as discussed in Sec. \ref{sec:cross_task_gen}. Unlike conventional methods that typically assume shared input-output spaces, LTSMs aim to transfer knowledge across heterogeneous time series datasets \cite{202}. Their strong generalization capability arises not from explicit OOD mechanisms, such as invariance enforcement, but rather from large-scale pre-training on diverse corpora, enabling them to capture universal representations of temporal dynamics applicable in zero-shot or few-shot scenarios. As illustrated in Fig. \ref{fig-llm}, LTSMs can be broadly divided into two main categories: (a) LLM-based methods, which leverage pre-trained LLMs for time series tasks, and (b) Time series foundation models, which are pre-trained from scratch on large time series data.

\input{table_llm}

\subsubsection{LLM-based Methods}

LLM-based methods adapt pre-trained language models to time series tasks by converting numerical sequences into token representations compatible with the LLM. While these approaches aim to harness the powerful sequence modeling capabilities demonstrated by LLMs on textual data, their effectiveness and necessity for numerical time series tasks, particularly forecasting, are subjects of ongoing debate. Recent critical evaluations, such as Tan et al. \cite{tan2024are}, suggest that several popular LLM-based forecasting methods may not offer significant performance improvements over simpler baselines or even models where the LLM component is removed. These studies indicate that observed gains might originate from other architectural elements like input encoders (e.g., patching), while the LLM itself introduces substantial computational overhead without commensurate benefits in standard forecasting benchmarks \cite{tan2024are}. Consequently, the value proposition of adapting LLMs designed for language to inherently different numerical sequence data remains an active research question. These methods can be further classified into fine-tuning/adapter-based approaches and prompt-based approaches.

Fine-tuning or adapter-based approaches preserve the general representation power of LLMs while adapting them to time series tasks through task-specific training or light-weight parameter updates. GPT4TS \cite{81} investigates leveraging pre-trained language models to build a universal framework for time series analysis, enabling strong cross-task generalization without domain-specific fine-tuning. Time-LLM \cite{70} introduces specialized input formats that convert time series into token sequences interpretable by LLMs, enhancing model adaptability. Moirai \cite{71} employs a unified training framework for diverse time series types, while LLM4TS \cite{72} aligns data distributions across multiple domains to reinforce robustness. ChatTime \cite{76} integrates multiple modalities into a unified time series foundation, allowing LLMs to handle a wide range of forecasting tasks. CALF \cite{75} applies cross-modal fine-tuning to connect numerical and textual information, further improving generalization across tasks. TransferKF \cite{68} investigates the use of language models for cardiovascular disease detection, demonstrating successful knowledge transfer from textual to temporal domains. ETP \cite{69} jointly pre-trains on ECG signals and textual data, enhancing the LLM’s ability to process both natural language and time series signals.

Prompt-based methods exploit the LLM’s inherent reasoning and knowledge capabilities without altering the model parameters. LLMTIME \cite{77} achieves competitive zero-shot forecasting by reformulating time series problems into textual prompts. TableTime \cite{80} transforms time series classification into table-understanding tasks, leveraging existing tabular reasoning capabilities to generalize across domains.

\subsubsection{Time Series Foundation Models}
Time series foundation models are designed and pre-trained specifically on time series corpora, learning universal temporal patterns directly from numerical data. Unlike LLM-based methods, these models do not rely on textual pre-training and instead capture domain-agnostic temporal dynamics from scratch. 

PatchTST \cite{67} uses an encoder-only architecture to learn time series representations for forecasting tasks. ForecastPFN \cite{65} demonstrates the effectiveness of large-scale synthetic pre-training. Jolt \cite{66} highlights challenges in achieving robust zero-shot generalization, emphasizing the need for architecture and data scale considerations. TimeGPT-1 \cite{79} pre-trains on multi-domain time series datasets, enabling robust cross-task transfer. TTMs \cite{73} employ a lightweight hybrid architecture combined with large-scale pre-training to achieve efficient zero-shot or few-shot multivariate time series forecasting while maintaining high accuracy. GTT \cite{144} is an encoder-only model designed for zero-shot multivariate forecasting, demonstrating strong generalization to OOD data.  CloudOps \cite{78} pretrains on cloud operations time series, demonstrating robust predictions under OOD conditions such as sudden anomalies or system changes. MOMENT \cite{83} is pre-trained on multi-domain data for forecasting, classification, and anomaly detection tasks. Lag-Llama \cite{82} extends time series modeling by integrating autoregressive and sequence-modeling capabilities. Timer \cite{84} explores GPT-style transformers for temporal modeling, while TimesFM \cite{85} adopts a decoder-only architecture optimized for computational efficiency and adaptability to non-stationary series. Chronos \cite{86} introduces the concept of a time series language, learning universal temporal patterns to enable generalization across diverse domains.

%% file: table_methods.tex
\begin{table*}
\centering
\caption{Comparison of decoupling-based, invariant-based, and adaptive mechanism-based classification methods.}
\label{tab:tsmethods}
\scriptsize
\begin{tabularx}{\textwidth}{%
  >{\RaggedRight\arraybackslash}p{1.5cm}  
  >{\RaggedRight\arraybackslash}p{3.2cm}  
  >{\RaggedRight\arraybackslash}X  
  >{\RaggedRight\arraybackslash}p{4cm}  
}
\toprule
Category&Subcategory & Model& Data Type\\
\midrule
\multirow{7}{*}{\makecell[{{p{1.5cm}}}]{Decoupling-based}}
 & Multi-Structured Analysis & ITSR \cite{34} & Multivariate discrete time series \\
 & Multi-Structured Analysis & SCNN \cite{35}, FSD \cite{36}, St-RKM \cite{37}, MSGNet \cite{40}, JointPGM  \cite{He2025TNNLS}, MetaEformer \cite{Huang2025MetaEformer}, FEDformer \cite{zhou2022fedformer}   & Multivariate time series \\
 & Multi-Structured Analysis & \(\beta\)-VAE \cite{38} & Timestamped time series \\
 & Causality-Inspired & Causal-HMM \cite{41}, CTSDG \cite{42}, LNN \cite{44} & Multivariate time series \\
 & Causality-Inspired & MaSDEs \cite{43} & Multivariate time series with noise  \\
 & Causality-Inspired & CCR \cite{45} & Perturbed univariate time series  \\
 & Causality-Inspired & DIOSC \cite{33},   \cite{Li2024TPAMI}, DIDA \cite{Zhang2022NeurIPS}, DeCau \cite{Liu2025tkde} & Cross-domain time series \\
\midrule
\multirow{8}{*}{\makecell[{{p{1.5cm}}}]{Invariant-based}}
 & Invariant Risk Minimization& FOIL \cite{46}, TSMixer \cite{50}, Meta-Learning \cite{51},TFPS \cite{sun2025learningpatternspecificexpertstime}  & Multivariate time series \\
  & Invariant Risk Minimization&Cedar \cite{Deng2024} & Continuous time series\\
 &  Invariant Risk Minimization& ROBUST \cite{47}, OOD-TV-IRM \cite{207}  & Multivariate discrete time series  \\
 & Invariant Risk Minimization & TSModel \cite{48} & Cross-domain time series  \\
 &   Domain-Invariance& Two-stage \cite{24}, TKNets \cite{25}, Diversify \cite{53}, In-N-Out \cite{56}, INNT \cite{dengalleviating}, \cite{Ott2022MM}, ST-Norm \cite{Hu2023icdm}, TCVAE \cite{He2024TNNLS}   & Multivariate discrete time series \\
 &  Domain-Invariance& Connect Later \cite{28},  PE-Att \cite{87} & Multivariate time series  \\
 & Domain-Invariance & Koodos \cite{52}, OODRL \cite{55} & Continuous time series  \\
 & Domain-Invariance & SMORE \cite{57}, EvoS \cite{58} & Discrete time series \\
\midrule
\multirow{7}{*}{\makecell[{{p{1.5cm}}}]{Adaptive Mechanism-based}}
 &  Explicit & CEP \cite{zhan2025continuousevolutionpooltaming},  \cite{oliveira2025dynamic}, \cite{Isaac2024},  CTOT \cite{26}, GradExp \cite{Stefanos2022ICML}, PSO \cite{Oliveira2017} & Univariate time series   \\
 &  Explicit &  PROCEED \cite{zhao2025kdd}, BTOA \cite{zhang2025batch},  HTSF \cite{Duan2023},  DoubleAdapt  \cite{49}, SOLID \cite{Chen2024KDD}, OneNet \cite{Zhang2023OneNet}& Multivariate time series  \\
  &  Explicit &ST-SSDL \cite{gao2025different} & Spatio-Temporal Time Series   \\
 &  Implicit &  Cogra \cite{Miyaguchi2019aaai}, TS\_Adam \cite{dong5570331rethinking}, Meta-Learning  \cite{51}& Univariate time series   \\
  &  Implicit &  HSN-LSTM \cite{Zheng2022tkde}& Multivariate time series  \\
 &  Implicit & WormKAN \cite{xu2024kandrift}, CORAL \cite{xu2025coralconceptdriftrepresentation}& Co-evolving time series \\
  &Implicit & FAD \cite{89} & Cross-domain time series  \\
\bottomrule
\end{tabularx}
\end{table*}

%% file: table_llm.tex
\begin{table*}[ht]
\centering
\setlength{\tabcolsep}{2pt}
\caption{Comparison between three classification methods for large-scale modeling of time series. Models are grouped into two principal categories: LLM-based methods that adapt LLMs to time series tasks (either via prompting or via fine-tuning / adapter tuning), and time series foundation models that are trained from scratch or primarily pre-trained on large time series corpora. Missing Support refers to the ability to handle missing data by imputing or preprocessing it before feeding it into a model. Flexible Distribution is a qualitative classification rather than a quantitative measure, focusing on probabilistic predictions rather than just point estimates. “-” indicates unknown. E, D, and E-D denote encoder-only, decoder-only, and encoder–decoder architectures, respectively.}
\begin{tabular}{>{\centering\arraybackslash}p{2.6cm}>{\centering\arraybackslash}p{2.5cm}>{\centering\arraybackslash}p{1.8cm}>{\centering\arraybackslash}p{1.0cm}>{\centering\arraybackslash}p{1.8cm}>{\centering\arraybackslash}p{0.7cm}>{\centering\arraybackslash}p{1.2cm}>{\centering\arraybackslash}p{1.7cm}>{\centering\arraybackslash}p{1.3cm}>{\centering\arraybackslash}p{1.4cm}}
\toprule
Category & Model & Venue & Open-source & Architec-tures & Zero-shot & Missing Support & Flexible Distribution & Trainable Token & Trainable Parameter \\
\midrule

\multirow{11}{*}{\parbox{2.6cm}{\centering LLM-based methods}} 
& GPT4TS \cite{81} & NeurIPS '23 & \textcolor{green}{\ding{51}} & E & \textcolor{green}{\ding{51}} & \textcolor{green}{\ding{51}} & \textcolor{green}{\ding{51}} & - & 84M \\
& TransferKF \cite{68} & EACL '23 & \textcolor{green}{\ding{51}} & E & \textcolor{green}{\ding{51}} & \textcolor{red}{\ding{55}} & \textcolor{red}{\ding{55}} & - & - \\
& LLMTIME \cite{77} & NeurIPS '23 & \textcolor{green}{\ding{51}} & E & \textcolor{green}{\ding{51}} & \textcolor{green}{\ding{51}} & \textcolor{green}{\ding{51}} & - & - \\
& Time-LLM \cite{70} & ICLR '24 & \textcolor{green}{\ding{51}} & E-D & \textcolor{green}{\ding{51}} & \textcolor{red}{\ding{55}} & \textcolor{red}{\ding{55}} & 7B & - \\
& Moirai \cite{71} & ICML '24 & \textcolor{green}{\ding{51}} & E & \textcolor{green}{\ding{51}} & \textcolor{green}{\ding{51}} & \textcolor{green}{\ding{51}} & 150B & 300M \\
& LLM4TS \cite{72} & NeurIPS '24 & \textcolor{red}{\ding{55}} & E & \textcolor{green}{\ding{51}} & \textcolor{red}{\ding{55}} & - & 3.4M & - \\
& ETP \cite{69} & ICASSP '24 & \textcolor{red}{\ding{55}} & E & \textcolor{green}{\ding{51}} & \textcolor{red}{\ding{55}} & \textcolor{red}{\ding{55}} & - & - \\
& ChatTime \cite{76} & AAAI '25 & \textcolor{green}{\ding{51}} & E & \textcolor{green}{\ding{51}} & \textcolor{green}{\ding{51}} & \textcolor{red}{\ding{55}} & 1B & 350M \\
& TableTime \cite{80} & CIKM '25 & \textcolor{green}{\ding{51}} & E & \textcolor{green}{\ding{51}} & \textcolor{red}{\ding{55}} & \textcolor{red}{\ding{55}} & - & - \\
& CALF \cite{75} & AAAI '25 & \textcolor{green}{\ding{51}} & E-D & \textcolor{green}{\ding{51}} & \textcolor{red}{\ding{55}} & \textcolor{red}{\ding{55}} & - & - \\
\midrule
\multirow{12}{*}{\parbox{2.6cm}{\centering Time series foundation models}}
& PatchTST \cite{67} & ICLR '23 & \textcolor{green}{\ding{51}} & E & \textcolor{red}{\ding{55}} & \textcolor{red}{\ding{55}} & \textcolor{red}{\ding{55}} & 20M & - \\
& ForecastPFN \cite{65} & NeurIPS '23 & \textcolor{green}{\ding{51}} & E & \textcolor{green}{\ding{51}} & \textcolor{green}{\ding{51}} & \textcolor{green}{\ding{51}} & - & - \\
& Jolt \cite{66} & NeurIPS '23 & \textcolor{red}{\ding{55}} & E & \textcolor{red}{\ding{55}} & \textcolor{red}{\ding{55}} & \textcolor{red}{\ding{55}} & - & - \\
& CloudOps \cite{78} & ArXiv '23 & \textcolor{green}{\ding{51}} & E-D & \textcolor{green}{\ding{51}} & \textcolor{red}{\ding{55}} & \textcolor{red}{\ding{55}} & - & 85.1M \\
& TimeGPT-1 \cite{79} & ArXiv '23 & \textcolor{red}{\ding{55}} & E-D & \textcolor{green}{\ding{51}} & \textcolor{green}{\ding{51}} & \textcolor{red}{\ding{55}} & 100B & - \\
& Lag-Llama \cite{82} & NeurIPS '23 &\textcolor{green}{\ding{51}}  & D &\textcolor{green}{\ding{51}}  &\textcolor{red}{\ding{55}} &\textcolor{red}{\ding{55}} & - & 3M \\
& GTT \cite{144} & CIKM '24 & \textcolor{green}{\ding{51}} & E & \textcolor{green}{\ding{51}} & \textcolor{green}{\ding{51}} & \textcolor{red}{\ding{55}} & 200M & 7M \\
& MOMENT \cite{83} & ICML '24 & \textcolor{green}{\ding{51}} & E & \textcolor{green}{\ding{51}} & \textcolor{green}{\ding{51}} & \textcolor{green}{\ding{51}} & 100B & 300M \\
& Timer \cite{84} & ICML '24 & \textcolor{green}{\ding{51}} & D & \textcolor{green}{\ding{51}} & \textcolor{green}{\ding{51}} & \textcolor{green}{\ding{51}} & 50B & 50M \\
& TimesFM \cite{85} & ICML '24 & \textcolor{green}{\ding{51}} & D & \textcolor{green}{\ding{51}} & \textcolor{red}{\ding{55}} & \textcolor{red}{\ding{55}} & 3T & 200M \\
& Chronos \cite{86} & TMLR '24 & \textcolor{green}{\ding{51}} & E-D & \textcolor{green}{\ding{51}} & \textcolor{green}{\ding{51}} & \textcolor{green}{\ding{51}} & 25B & 700M \\
& TTMs \cite{73} & NeurIPS '24 & \textcolor{green}{\ding{51}} & D & \textcolor{green}{\ding{51}} & \textcolor{red}{\ding{55}} & - & 5M & - \\
& TimeRAF \cite{11031238} & TKDE '25 & \textcolor{green}{\ding{55}} & E & \textcolor{green}{\ding{51}} & \textcolor{red}{\ding{55}} &  \textcolor{red}{\ding{55}}& - & 1M \\

\bottomrule
\end{tabular}
\label{tab:models_summary}
\end{table*}

%% file: 6_sec_ood_evaluation.tex
OOD evaluation aims to quantify a model’s generalization ability under unknown data distributions. Based on the completeness of the evaluation data, existing approaches can be divided into two categories: extrinsic evaluation and intrinsic evaluation. We summarize their differences from OOD assessment in Table \ref{ood_evaluation}. These two complementary approaches form a comprehensive evaluation framework spanning data to model.

\input{table_evaluation}

\subsection{Extrinsic OOD Evaluation} 
Extrinsic evaluation constructs or gathers explicit OOD test data to assess the model’s generalization performance directly. According to the source and labeling status of the test data, extrinsic evaluation can be further subdivided into verifiable and speculative assessments. Table \ref{evaluation} provides a detailed breakdown of the categories of Extrinsic OOD Evaluation and their corresponding datasets.

\subsubsection{Verifiable OOD Evaluation} 
Verifiable OOD evaluation measures a model’s OOD generalization performance on labeled test datasets across different time points. Although assessing models on OOD datasets seems intuitive, ensuring reliable and reproducible evaluation outcomes requires the design of rigorous protocols. These protocols should specify how distribution shifts are induced, how data environments are partitioned, and how fairness is maintained under different testing conditions. For instance, Meng et al. \cite{22,23} employ stable explanation methods to validate model performance in various data environments and provide quantitative metrics and visualizations, enhancing transparency and verifiability in the evaluation process. In addition, some researchers \cite{41,30} introduce causal modeling and information bottleneck strategies, which help reveal how latent factors affect a model’s OOD performance, thereby adding interpretability to the evaluation. Furthermore, drawing on research on capturing spatiotemporal invariance and multi-scale correlations \cite{64,35,40} offers deeper insights into how models behave under real-world distribution shifts.

\subsubsection{Speculative OOD Evaluation} 

In real-world scenarios, obtaining labeled OOD test data is often highly challenging. Consequently, researchers typically rely on unlabeled test data to assess model applicability or identify the optimal model. In this context, speculative OOD evaluation emerges as an effective strategy. Various methods have been explored and adopted to mitigate the lack of labels for OOD test data, including synthetic data generation \cite{20}, pseudo-label inference \cite{56}, and dynamic test customization at the inference stage \cite{57}. Synthetic data is a common OOD evaluation strategy to simulate model behavior under complex conditions. For instance, Narwariya et al. \cite{87} leverage the PowerTAC simulator to generate electricity consumption data under different pricing strategies, thus examining model robustness within dynamic market environments. Similarly, Yuan et al. \cite{24} employ domain-mapping techniques to convert desktop-based gesture data into virtual reality gesture data, thereby alleviating the shortage of labeled VR datasets and providing a new source for OOD evaluation. When labels are unavailable, pseudo-label generation offers another effective approach for OOD evaluation. Tie et al. \cite{56} propose the In-N-Out framework, which employs an auxiliary-information-guided self-training mechanism to produce pseudo-labels and optimize model adaptability. Dynamic inference strategies furnish real-time adaptability solutions for OOD evaluation. For example, Wang et al. \cite{57} develop the SMORE framework, which leverages hyperdimensional domain adaptation at inference time to adjust a multi-sensor time series classification model dynamically. The model can quickly adapt to sensor distribution shifts during testing by exploiting a similarity-based feature reorganization mechanism. Furthermore, liquid neural networks \cite{44}, which can modify their parameters in real-time, complement generative approaches to form a complete pipeline.

\subsection{Intrinsic OOD Evaluation} 
Intrinsic OOD evaluation refers to inferring a model's potential generalization capability by analyzing its intrinsic properties when test datasets are unavailable. Current research has extensively investigated critical characteristics, including distributional robustness \cite{53,55,45}, smoothness \cite{52,89}, and invariance \cite{46,42}, which fundamentally govern model generalization and enhance performance on unseen data. Distributional robustness ensures stable and reliable predictions under distribution shifts. For instance, Tonin et al. \cite{37} validated the effectiveness of distributional robustness-based methods through similarity metrics and energy function perspectives, respectively. Meanwhile, Lu et al. \cite{54} established a unified framework to directly quantify model responses to OOD data, while \cite{28} demonstrated enhanced adaptability to distribution shifts via targeted data augmentation. For example, \cite{52} proposed continuous temporal domain generalization by capturing the smooth temporal evolution of model outputs to predict their performance on unseen data, whereas \cite{58} further explored how normalization strategies achieve sustained smooth adaptation against temporal drift. Invariance-oriented approaches focus on extracting core features insensitive to external disturbances, thereby maintaining high performance under OOD scenarios. Specifically, \cite{42} and \cite{45} employed causal inference and causal representation learning to disentangle confounding factors, while \cite{33} utilized contrastive learning for feature decoupling to reinforce invariant internal representations. Notably, although more challenging in the absence of test data, this intrinsic evaluation paradigm provides deeper insights into the essence of OOD generalization. It compels models to concentrate on relevant features while maintaining stable and accurate predictions across diverse OOD scenarios.

%% file: table_evaluation.tex
\begin{table}
\centering
\caption{Comparison of Extrinsic and Intrinsic OOD Evaluation. Comparison of Extrinsic and Intrinsic OOD Evaluation. The green check mark (\textcolor{green}{\ding{51}}) indicates that the specified data or labels are available, while the red mark (\textcolor{red}{\ding{55}}) indicates they are unavailable.}
\label{ood_evaluation}
\setlength{\tabcolsep}{1pt} 
\begin{tabular}{>{\centering\arraybackslash}p{2.3cm}>{\centering\arraybackslash}p{1.6cm}>{\centering\arraybackslash}p{1.6cm}>{\centering\arraybackslash}p{2.2cm}}
\toprule
\multirow{2}{*}{Dimension} &  \multicolumn{2}{c}{Extrinsic OOD Evaluation} &   \multirow{2}{*}{\centering \makecell{Intrinsic \\ OOD Evaluation}}  \\  \cmidrule(lr){2-3}
                     & Verifiable & Speculative &   \\ 
\midrule
OOD test data    & \textcolor{green}{\ding{51}}  & \textcolor{green}{\ding{51}}  &   \textcolor{red}{\ding{55}}  \\ 
Label availability  &\textcolor{green}{\ding{51}}  &  \textcolor{red}{\ding{55}}  &    -  \\ 
Evaluation basis & True labels  & Uncertainty  & Internal features  \\ 
\bottomrule
\end{tabular}
\end{table}

\begin{table*}
\centering
\caption{Extrinsic OOD Evaluation Type.}
\scriptsize
\label{evaluation}
\begin{tabularx}{\linewidth}{ 
  >{\RaggedRight\arraybackslash}p{1cm}
  >{\RaggedRight\arraybackslash}p{2.3cm}
  >{\RaggedRight\arraybackslash}X
  >{\RaggedRight\arraybackslash}p{1cm}
  >{\RaggedRight\arraybackslash}p{2.1cm}
  >{\RaggedRight\arraybackslash}p{3cm}
}
\toprule
Types & Methods & Datasets & Types & Methods & Datasets \\
\midrule
Verifiable& SLIME-MTS \cite{22} &  \textcolor{blue}{\href{https://www.timeseriesclassification.com/index.php}{UEA}} & Verifiable & TCVAE \cite{He2024TNNLS} &   \textcolor{blue}{\href{https://archive.ics.uci.edu/dataset/204/pems+sf}{PEMS-SF}; \textcolor{blue}{\href{https://archive.ics.uci.edu/dataset/321/electricityloaddiagrams20112014}{Electricity}}; \textcolor{blue}{\href{https://www.nrel.gov/grid/solar-power-data}{Solar-Energy}}; \textcolor{blue}{\href{https://github.com/CSSEGISandData/COVID-19/tree/master}{COVID-19}}}; \textcolor{blue}{\href{https://pems.dot.ca.gov/}{PeMSD7(M)}}; \textcolor{blue}{\href{https://github.com/liyaguang/DCRNN}{METR-LA}}\\
Verifiable & ST-Norm \cite{Hu2023icdm}& \textcolor{blue}{\href{https://drive.google.com/drive/folders/10FOTa6HXPqX8Pf5WRoRwcFnW9BrNZEIX}{METR-LA}}; \textcolor{blue}{\href{https://archive.ics.uci.edu/dataset/501/beijing+multi+site+air+quality+data}{Beijing Air Quality}} &Verifiable& INNT \cite{dengalleviating} &  \textcolor{blue}{\href{https://drive.google.com/drive/folders/1ZOYpTUa82_jCcxIdTmyr0LXQfvaM9vIy}{ETT; Weather; Traffic}}; \textcolor{blue}{\href{https://drive.google.com/drive/folders/1ZOYpTUa82_jCcxIdTmyr0LXQfvaM9vIy}{Electricity}} \\
Verifiable & Causal-HMM \cite{41} &  \textcolor{blue}{\href{https://github.com/LilJing/causal_hmm}{In-house data on PPA}} &  Verifiable &  DIDA  \cite{Zhang2022NeurIPS} &  \textcolor{blue}{\href{https://drive.google.com/file/d/19SOqzYEKvkna6DKd74gcJ50Wd4phOHr3/view?usp=share_link}{COLLAB}};\textcolor{blue}{\href{https://business.yelp.com/data/resources/open-dataset/}{Yelp}};\textcolor{blue}{\href{https://drive.google.com/file/d/19SOqzYEKvkna6DKd74gcJ50Wd4phOHr3/view?usp=share_link}{Transaction}}\\
Verifiable & TimeX++ \cite{30} & \textcolor{blue}{\href{https://www.cs.ucr.edu/~eamonn/time_series_data_2018}{UCR}} & Speculative & PE-Att \cite{87} & Generating simulated data\\
Verifiable & CauSTG \cite{64} &  \textcolor{blue}{\href{https://data.cic-tp.com/h5/sample-data/china/export-data/company/suzhou-industrial-park}{SIP}}; 
\textcolor{blue}{\href{https://www.kaggle.com/datasets/annnnguyen/metr-la-dataset}{METR-LA}}; 
\textcolor{blue}{\href{https://github.com/shuowang-ai/PM2.5-GNN}{KnowAir}}; 
\textcolor{blue}{\href{https://github.com/laiguokun/multivariate-time-series-data/tree/master/electricity}{Electricity}} 
& Speculative & LNN \cite{44} & Generating data \\
Verifiable & SCNN \cite{35} & \textcolor{blue}{\href{https://github.com/laiguokun/multivariate-time-series-data}{Traffic}}; 
\textcolor{blue}{\href{https://github.com/laiguokun/multivariate-time-series-data}{Solar-energy}}; 
\textcolor{blue}{\href{https://github.com/laiguokun/multivariate-time-series-data}{Electricity}}; 
\textcolor{blue}{\href{https://github.com/laiguokun/multivariate-time-series-data}{Exchange-rate}} 
& Speculative & VRStrokeOOD\cite{24} & Generating unlabeled data \\
Verifiable & MSGNet \cite{40} & \textcolor{blue}{\href{https://opensky-network.org}{Flight}}; 
\textcolor{blue}{\href{https://drive.google.com/drive/folders/1ZOYpTUa82_jCcxIdTmyr0LXQfvaM9vIy}{Weather}}; 
\textcolor{blue}{\href{https://drive.google.com/drive/folders/1ZOYpTUa82_jCcxIdTmyr0LXQfvaM9vIy}{ETT}}; 
\textcolor{blue}{\href{https://github.com/laiguokun/multivariate-time-series-data/tree/master/electricity}{Exchange-Rate}}; 
\textcolor{blue}{\href{https://github.com/laiguokun/multivariate-time-series-data/tree/master/electricity}{Electricity}} 
& Speculative & ACGAN \cite{86} & Unknown attack data \\
Verifiable & MetaEformer \cite{Huang2025MetaEformer} & \textcolor{blue}{\href{https://github.com/EdgeBigBang/KDD25_MetaEformer/tree/main/dataset}{CBW}};\textcolor{blue}{\href{https://github.com/EdgeBigBang/KDD25_MetaEformer/tree/main/dataset}{ECW}};\textcolor{blue}{\href{https://github.com/EdgeBigBang/KDD25_MetaEformer/tree/main/dataset}{ECL}};\textcolor{blue}{\href{https://github.com/EdgeBigBang/KDD25_MetaEformer/tree/main/dataset}{Traffic}} & Speculative& SBO \cite{20} & Generating unlabeled data   \\
Verifiable &  FEDformer \cite{zhou2022fedformer} & \textcolor{blue}{\href{https://drive.google.com/drive/folders/1ZOYpTUa82_jCcxIdTmyr0LXQfvaM9vIy}{ETT}};  \textcolor{blue}{\href{https://drive.google.com/drive/folders/1ZOYpTUa82_jCcxIdTmyr0LXQfvaM9vIy}{Electricity}}; \textcolor{blue}{\href{https://drive.google.com/drive/folders/1ZOYpTUa82_jCcxIdTmyr0LXQfvaM9vIy}{Exchange}};  \textcolor{blue}{\href{https://drive.google.com/drive/folders/1ZOYpTUa82_jCcxIdTmyr0LXQfvaM9vIy}{Traffic}}; \textcolor{blue}{\href{https://drive.google.com/drive/folders/1ZOYpTUa82_jCcxIdTmyr0LXQfvaM9vIy}{ILI}}  & Speculative & In-N-Out \cite{56} & Generating pseudo-labels \\
Verifiable & DeCau \cite{Liu2025tkde} &   \textcolor{blue}{\href{https://www.nyc.gov/site/tlc/about/tlc-trip-record-data.page}{NYC Taxi}}; \textcolor{blue}{\href{https://www.microsoft.com/en-us/research/publication/t-drive-trajectory-data-sample/}{Beijing Trajectory}} &Speculative & SMORE \cite{57} & Unlabeled USC-HAD\\
Verifiable & Cedar \cite{Deng2024}& \textcolor{blue}{\href{https://www.kaggle.com/c/favorita-grocery-sales-forecasting}{Favorita}}; \textcolor{blue}{\href{https://www.fhwa.dot.gov/policyinformation/tables/tmasdata/}{US-traffic}};  \textcolor{blue}{\href{https://www.kaggle.com/datasets/mattiuzc/stock-exchange-data?select=indexProcessed.csv}{Stock-volume}}& Speculative &Cedar \cite{Deng2024} & Generating data    \\
\bottomrule
\end{tabularx}
\begin{tablenotes}
\footnotesize
\item Note: For methods using duplicate datasets, only representative datasets are listed.
\end{tablenotes}
\end{table*}

%% file: 7_sec_applications.tex
This section explores applications of TS-OOG across multiple domains, highlighting the urgent need for high generalization capacity in each, as shown in Fig. \ref{fig-app}. As data collection environments become increasingly complex, data from various domains often exhibit distribution shifts due to external factors. Consequently, developing robust and adaptable models across domains has become paramount.

\textit{1) Transportation.} In the transportation, autonomous driving systems and smart city infrastructures rely on time series data from various sensors, such as cameras and radar. Data gathered by autonomous vehicles often undergo significant distribution shifts due to weather conditions, lighting variations, and road surface changes \cite{90,44}. Thus, Researchers focus on real-time correction of distributional shifts to ensure correct decision-making under extreme conditions \cite{146}. Meanwhile, the multitude of sensors used in smart cities (e.g., traffic flow monitoring, public facilities, environmental detection) generate data that, when effectively integrated via TS-OOG techniques, can substantially improve the intelligence of urban management \cite{91,48,92}. In transportation, demand for data distribution–agnostic solutions extends to real-time decision-making and large-scale data integration and management.

\textit{2) Environment.} Air quality data often display dynamic characteristics influenced by seasonal variations, pollution events, and meteorological fluctuations \cite{93}. In contrast, remote sensing data exhibit inconsistencies due to differences in collection devices, observation angles, and geographic locations \cite{101}. Likewise, energy consumption data are affected by various factors such as economic activity, climate variability, and policy interventions, making their time series patterns complex and volatile \cite{87}. To effectively capture these evolving dynamics, researchers employ TS-OOG methods to model data trends, enhancing predictive accuracy and stability in environmental monitoring and management \cite{94}. This highlights the importance of strong generalization capacity in addressing the challenges of highly variable environmental data.

\begin{figure}
	\centering
	\resizebox{.5\textwidth}{!}{
	\input{fig_application}
	}
	\caption{Several applications of TS-OOG.}
	\label{fig-app}
\end{figure}

\textit{3) Public Health.}
TS-OOG also proves advantageous in public health. During AI-assisted drug discovery, abundant molecular dynamics simulations, experimental data, and clinical trial results undergo distribution shifts due to varying experimental conditions and individual patient differences. TS-OOG methodologies enable models to maintain high consistency and predictive accuracy across these diverse experimental settings, expediting drug development \cite{58}. Additionally, psychological health monitoring depends on physiological and behavioral data, which can differ considerably among different populations and contexts. Constructing diagnostic systems capable of handling distributional shifts allows earlier detection of potential mental health issues. For instance, Yang et al. \cite{95} integrate EEG and ECG data to achieve robust generalization in epilepsy detection. In public health, addressing distributional shifts is critical for accurate diagnosis and prediction.


\textit{4) Public Safety.}
System logs and network traffic data in cybersecurity often exhibit anomalous patterns under novel attacks. Moreover, human activity recognition systems frequently rely on temporal data collected from wearable or surveillance devices; such data are easily affected by individual variability and environmental factors, leading to distributional shifts \cite{96}. Researchers are, therefore, actively pursuing cross-user and cross-scenario modeling methods to achieve stable recognition. These applications demonstrate how OOD generalization methods can effectively mitigate challenges posed by data changes, thereby strengthening public safety.

\textit{5) Others.} 
It is worth noting that TS-OOG techniques exhibit wide-ranging potential in other areas. In finance, for instance, stock prices, trading volumes, and macroeconomic indicators are inherently non-stationary, heavily influenced by regulatory changes, market sentiment, and external economic environments. OOD generalization can help financial models adapt to these dynamic shifts, offering more reliable data support for market prediction, risk management, and anomaly detection \cite{97} Similarly, for industrial equipment fault diagnostics \cite{63}, OOD generalization shows promise in anomaly detection and preventive strategies. By continuously exploring and applying TS-OOG techniques across diverse domains, systems achieve higher robustness under extreme conditions and more trustworthy data-driven decision-making.

%% file: fig_application.tex
\begin{forest}
  for tree={
  grow=east,
  reversed=true,
  anchor=base west,
  parent anchor=east,
  child anchor=west,
  base=left,
  rectangle,
  draw,
  rounded corners,align=left,
  minimum width=2.5em,
  inner xsep=4pt,
  inner ysep=1pt,
  },
  where level=1{text width=5em,fill=blue!10}{},
  where level=2{text width=5em,font=\footnotesize,fill=pink!30}{},
  where level=3{font=\footnotesize,yshift=0.26pt,fill=yellow!20}{},
  [TS-OOG\\Applications,fill=green!20
    [Transportation,text width=5.8em
        [Autonomous driving,text width=7.4em
            [e.g. LNNs \cite{44}/  \cite{90} /  \cite{146}.
            ]
        ]
        [Smart city,text width=4em
          [e.g. TSModel \cite{48} /  \cite{91}  /  \cite{92}.
          ]
        ]
     ]
    [Environment,text width=5.2em
        [Air quality,text width=4em
            [e.g. DAQFF \cite{93}.
            ]
        ]
        [Remote sensing,text width=6em
         [e.g.   AQNet \cite{101}.
         ]
        ]
        [Energy consumption,text width=7.4em
            [e.g.  PE-Att  \cite{87} /  \cite{94}.
            ]
        ]
     ]
    [Public health,text width=5.2em
        [Drug discovery,text width=5.4em
            [e.g. EvoS \cite{58}.
            ]
        ]
        [Mental health,text width=5.2em
            [e.g.  EEG-ECG \cite{95}.
            ]
        ]
    ]
    [Public safety,text width=5em
        [Cybersecurity,text width=4.8em
            [e.g. AFFAR  \cite{96}.
            ]
        ]
    ]
    [Others,text width=2.8em
        [Robust,text width=3em
            [e.g. FOIL \cite{46}.
            ]
        ]
        [Finance,text width=3.5em
            [e.g. \cite{97}.
            ]
        ]
        [Fault diagnosis,text width=5.4em
            [e.g. Deep ensembles \cite{63}.
            ]
        ]
        [Human activity \\ recognition,text width=6em
            [e.g. DTE \cite{61}/  \cite{24}. 
            ]
        ]
    ]
]
\end{forest}

        

%% file: 8_sec_future.tex
This section discusses several challenges and future directions for TS-OOG.

\textit{1) Dynamic Distribution Shifts and Continual Adaptation Mechanisms.} In open and dynamic environments, machine learning systems should effectively handle continuous changes in data distributions, which primarily manifest as covariate shift and concept drift \cite{174,175}. These shifts lead to degradation in model prediction accuracy, where traditional batch training methods fail to capture gradual changes in data streams and may cause critical decision failures in high-stakes applications. Current research focuses on constructing cognitive closed-loop systems that enhance model adaptability through dynamic detection \cite{147,148}, continual adaptation \cite{149,150}, and trustworthiness verification \cite{63,151}. Future investigations should prioritize the optimization of online algorithms for efficient shift identification under constrained computational resources, establish task sequence knowledge evaluation frameworks to quantify catastrophic forgetting effects in continual learning and derive generalization error bounds in dynamic environments to provide theoretical foundations for system design. As machine learning systems progressively adapt to environmental feedback, they will achieve self-evolution and synergistic development in dynamic environments.

\textit{2) Constructing Robust Invariant Representations.} Time series data are often influenced by seasonality, trends, and abrupt events, making it challenging to capture cross-period and cross-domain invariant information \cite{98, 100}. Future research may focus on feature-separation strategies grounded in decoupling, causal inference, and the information bottleneck principle, thereby isolating stable patterns while adapting to distribution shifts \cite{179,180}. For instance, multi-scale decomposition can be used to model short-term cyclical fluctuations and long-term trends \cite{176}. At the same time, causal graphs combined with adversarial training can suppress non-essential confounding variables; both approaches help models adapt to distribution shifts. Developing invariant representations will provide a solid theoretical and methodological foundation to enhance generalization in unfamiliar environments.

\textit{3) Multimodal Unified LTSMs.} Existing time series models often struggle in zero-shot or few-shot scenarios with OOD generalization, encountering bottlenecks in cross-domain knowledge transfer, limited interpretability, and insufficient robustness \cite{106,107}. A promising direction is to combine large-scale multimodal pre-training, task reformulation, and model reprogramming to construct a unified large-scale time series modeling framework \cite{183}. Such a framework can deliver efficient generalization and in-depth analytics across finance, healthcare, and industrial domains by incorporating domain adaptation mechanisms. Developing a multimodal unified LTSM will establish a strong technical foundation for intelligent decision-making and automated systems in various industries, thereby driving a new wave of growth in artificial intelligence applications.

\textit{4) Uncertainty Quantification and Benchmark OOD Evaluation for Time Series.} Building a comprehensive, dynamic, and adaptive OOD evaluation system requires accurately quantifying model uncertainty while thoroughly assessing OOD performance in time series data \cite{104,105}. Key challenges include limited data coverage and evolving distributions. Insufficient data diversity compromises model generalization in noisy or extreme scenarios \cite{184}. In contrast, continuous changes in data and business requirements call for real-time monitoring and ongoing optimization of evaluation procedures. Moreover, the absence of standardized, interpretable evaluation metrics hinders effective model improvement, and difficulties associated with collecting regulatory-compliant. Possible solutions include constructing layered, multimodal datasets to enhance coverage and diversity, implementing real-time distribution monitoring and online learning for continual updates, and establishing interpretable multidimensional metrics to support diagnostic and visualization needs. Ultimately, we view the development of a unified framework that enables cross-platform and cross-model integration while combining uncertainty quantification with time series OOD evaluation as both necessary and feasible.

\textit{5) Explainable AI for TS-OOG.} Explainability is crucial for enhancing the transparency, trustworthiness, and accountability of AI systems, particularly in high-risk areas such as healthcare \cite{204} and autonomous driving \cite{205}. As AI applications expand, balancing model robustness and explainability in dynamic environments has become increasingly urgent \cite{102,103,186}. While deep learning models excel at capturing temporal dependencies and delivering high predictive accuracy, their decision-making processes often lack transparency in OOD generalization, undermining trust in high-stakes scenarios \cite{152,154}. Furthermore, to ensure transparent and secure decision-making, developers and domain experts must thoroughly examine and understand such models before deployment. Therefore, future research and applications should delve into these models' internal evolution and decision logic in dynamic time-series contexts while devising continual learning strategies that uphold performance and interpretability. We can foster the real-world implementation of explainable AI in complex and evolving environments by developing lightweight, robust time series analysis frameworks that integrate domain knowledge.